\documentclass[11pt]{article}

\usepackage[preprint]{acl}

\usepackage{times}
\usepackage{latexsym}
\usepackage{caption}

\usepackage[T1]{fontenc}

\usepackage[utf8]{inputenc}

\usepackage{microtype}

\usepackage{inconsolata}

\usepackage{graphicx}
\usepackage{amsmath} 
\usepackage{booktabs}
\usepackage{multirow}
\usepackage[most]{tcolorbox}

\usepackage{tabularx}
\usepackage{array}
\usepackage{makecell}
\usepackage{listings}
\usepackage{float}
\usepackage{caption}
\usepackage{amssymb}
\usepackage[inline]{enumitem}
\usepackage{subcaption}
\usepackage[table]{xcolor}
\usepackage{booktabs,amsmath}
\usepackage{marvosym}

\usepackage{xcolor,colortbl,booktabs,tabularx,array}
\definecolor{cSem}{RGB}{157,39,41}
\definecolor{cLex}{RGB}{187,146,72}
\definecolor{cSyn}{RGB}{124,87,124}
\definecolor{cStr}{RGB}{73,106,87}
\definecolor{cSty}{RGB}{90,113,145}
\newcommand{\simgrp}[2]{\cellcolor{#1!80}&\multicolumn{2}{>{\columncolor{#1!9}}l}{\textcolor{#1!85!black}{\textbf{\textit{#2}}}}\\}
\newcommand{\simrow}[1]{\cellcolor{#1!80}&\cellcolor{#1!3.5}}
\newcommand{\simcc}[1]{\cellcolor{#1!3.5}}
\newcommand{\simgap}{\arrayrulecolor{white}\specialrule{1.2pt}{0pt}{0pt}\arrayrulecolor{black}}

\definecolor{meaningbg}{HTML}{DCECF8}
\definecolor{formbg}{HTML}{F8E9D7}
\definecolor{nbar}{HTML}{7D8590}
\definecolor{ntrack}{HTML}{E6E8EB}
\newlength{\nbarw}
\newcommand{\nv}[1]{#1\hspace{2.5pt}\rlap{\color{ntrack}\rule[0.25ex]{\nbarw}{1.15ex}}{\color{nbar}\rule[0.25ex]{#1\nbarw}{1.15ex}}\hspace{\dimexpr\nbarw-#1\nbarw\relax}}
\newcommand{\hd}[1]{\multicolumn{1}{r}{\textbf{#1}}}

\title{Revisiting LLM Homogeneity Through a Multidimensional Lens}

\author{
  Ziliang Qiu \\
  University of Illinois Urbana-Champaign
  \And
  Renfen Hu\textsuperscript{\Letter} \\
  Beijing Normal University \\
  \texttt{irishu@mail.bnu.edu.cn}
  \AND
  Yi Ding \\
  Purdue University
  \And
  Ted Underwood \\
  University of Illinois Urbana-Champaign
}

\begin{document}
\maketitle
\begin{abstract}
Concerns about LLM homogeneity are typically quantified with embedding similarity, which reduces two texts to a single score. We revisit homogeneity with 25 interpretable metrics spanning five dimensions: semantic, lexical, syntactic, structural, and stylistic \& presentational. A human study shows that perceived similarity depends on both content (what responses say) and form (how they say it), yet embedding similarity correlates less with human meaning judgments than sentence-level matching ($\rho=0.417$ vs.\ $0.606$) and not significantly with form judgments ($\rho=0.140$). Across 18 contemporary LLMs, models say similar things in different words: they share global semantics and broad linguistic profiles but remain distinct in lexical characteristics, structural patterns, and interactional conventions. Across releases from 2024 to 2026, models become more like themselves but not more like one another: among significantly changing metrics, intra-model distances decrease for 17 of 22, and inter-model distances increase for 14 of 21. Across two post-training pipelines, model-specific conventions emerge during SFT and are selectively reshaped by preference optimization. These results suggest that homogeneity is multidimensional and evolving, and that embedding similarity alone can overstate cross-model homogeneity while concealing diverging intra- and inter-model trends.\footnote{Our code and data are publicly available at \url{https://github.com/ziliang6/HomogeneityEvaluation}}
\end{abstract}

\section{Introduction}

\begin{figure}[!t]
    \centering
    \includegraphics[width=\columnwidth]{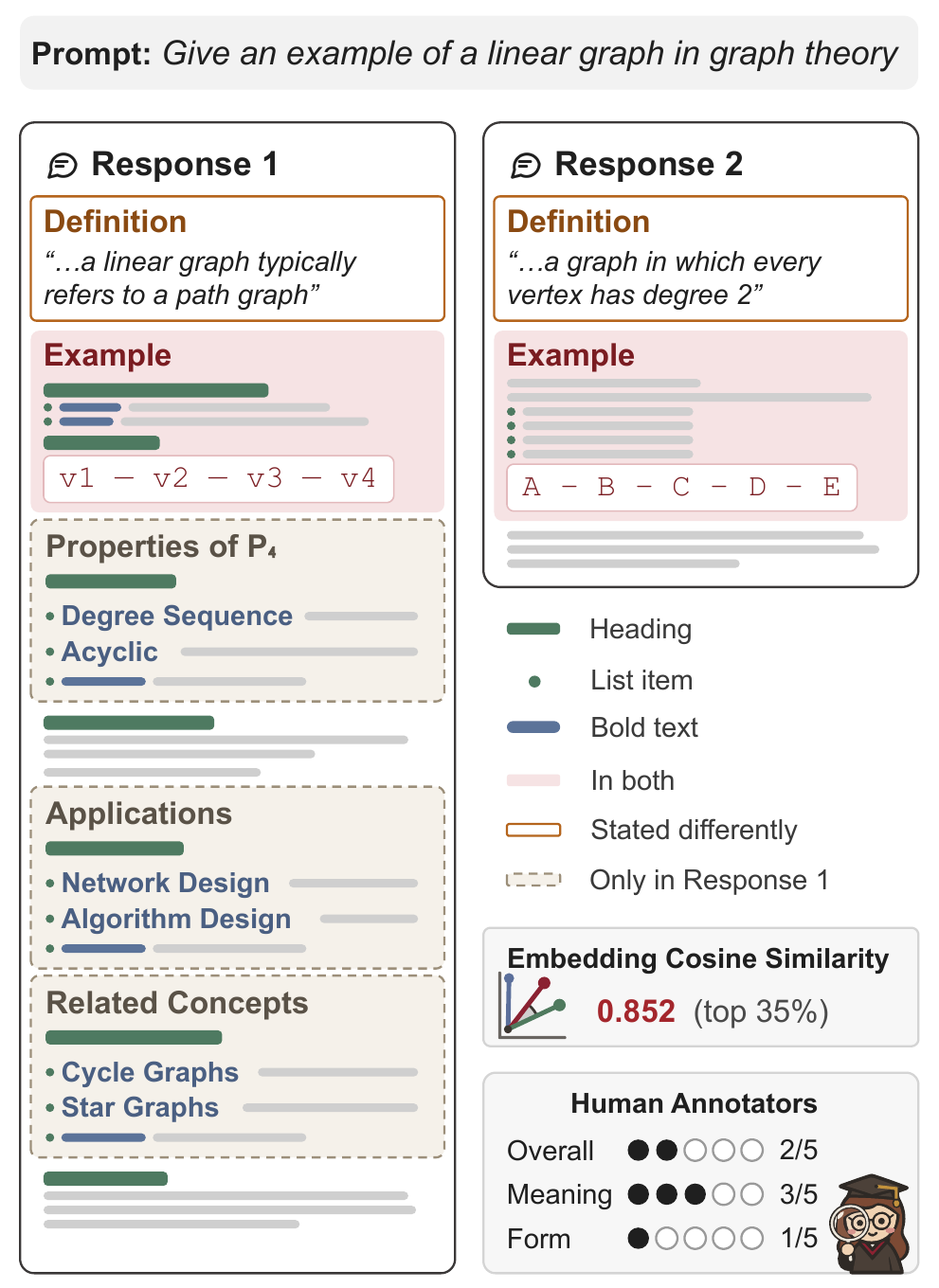}
    \caption{An example pair from \textsc{Infinity-Chat}~\citep{jiang2025artificial} showing embedding similarity (Qwen3-Embedding-8B) and human similarity ratings (\S\ref{sec:human_similarity}). The responses differ in definition and structure. Despite high embedding similarity, annotators rate the pair 2/5 overall, 3/5 for meaning, and 1/5 for form. See Figure~\ref{fig:case-study-responses} for the full example.}
    \label{fig:motivating-example}\vspace{-1em}
\end{figure}

Text embeddings have been widely used to measure semantic similarity between texts \citep{reimers2019sentence,karpukhin2020dense,lewis2020retrieval}. More recently, motivated by concerns about model collapse and declining output diversity \citep{shumailov2024ai,seddik2024bad,padmakumar2024does}, embedding-based similarity has also been adopted to quantify LLM response homogeneity \citep{kirk2024understanding,guo2025benchmarking,wenger2025we,jiang2025artificial}. 

However, embedding similarity provides only a partial view of textual similarity.
Modern general-purpose embedding models are mainly trained for semantic matching and retrieval \citep{li2023towards,xiao2024c,qwen3embedding}, emphasizing topical relatedness and shared themes. Consequently, high embedding similarity generally signals semantic alignment, but does not guarantee similarity in specific content. 
This limitation becomes especially critical in evaluating LLM responses, as answers to the same prompt often intrinsically share topic and task context \citep{schaeffer2026strong}. Moreover, two responses can share meaning (\emph{what} is said) but differ significantly in \emph{how} they convey it, using varied vocabulary, syntax, or organization. Such differences in expression are likewise difficult for embedding similarity to capture.
As shown in Figure~\ref{fig:motivating-example}, the two responses not only cover different content but also differ substantially in length and presentation, yet still receive a high embedding similarity score (0.852). Human annotators, however, rate them as only moderately similar in meaning and highly dissimilar in form.

Furthermore, a single embedding score offers limited interpretability~\citep{zhong2022towards,leiter2024towards}, providing little insight into which textual properties are shared or distinct~\citep{yang2026structsimmeasuringideasimilarity}. This limitation matters because influential conclusions about LLM homogeneity rest on such scores: studies find that LLMs converge toward an \emph{Artificial Hivemind}~\citep{jiang2025artificial}, and that this convergence has intensified over time~\citep{patel2026llms}. We ask whether these conclusions still hold once content and form are measured separately, which requires a multidimensional and interpretable account of textual similarity.

In this study, we propose a multidimensional framework for textual similarity that incorporates 25 interpretable metrics spanning five dimensions: semantic, lexical, syntactic, structural, and stylistic \& presentational.
Using this framework, we first examine how humans perceive textual similarity, what embeddings capture, and what they miss.
Our findings show that both meaning and form contribute to human judgments, while whole-response embedding similarity shows a lower observed correlation with perceived meaning than sentence-level matching, and largely overlooks structural features important to form judgments (\S~\ref{sec:human_embedding_check}).

Further, we systematically analyze LLM response homogeneity from both synchronic (cross-model) and diachronic (evolutionary and post-training) perspectives:

\textbf{Cross-model:} Embedding-based similarity yields the highest normalized inter-model similarity among all 25 metrics ($N=0.905$), substantially exceeding lexical-overlap measures such as BLEU and ROUGE-L ($N=0.341$ and $0.505$), which indicate far less overlap in wording. Meanwhile, we find models are similar in several grammatical and affective metrics, but show pronounced differences in lexical characteristics, structural patterns, and interactional conventions—a pattern that holds across diverse prompt types (\S~\ref{sec:current}).

\textbf{Cross-generation:} Models become more self-consistent in their responses, yet differences between models in fine-grained content, wording, and interactional conventions remain—and, in some aspects, even increase—despite greater convergence in global semantic representations (\S~\ref{sec:temporal}).

\textbf{Cross-stage:} Model-specific response conventions emerge during supervised fine-tuning and are selectively reshaped in subsequent training phases. While later stages generally increase semantic similarity, they also amplify or reduce distinctions in structure, wording, and formatting along complex, dimension-specific trajectories (\S~\ref{sec:stage}).

Together, these findings show that the \emph{Artificial Hivemind} has two facets that evolve in opposite directions: intra-model repetition intensifies, while inter-model homogeneity weakens along many textual dimensions, a divergence that whole-response embeddings conceal.

\begin{table*}[t]
\centering
\begin{NoHyper} 
\fontsize{9}{10.8}\selectfont
\setlength{\tabcolsep}{4pt}
\renewcommand{\arraystretch}{1.15}
\setlength{\aboverulesep}{0pt}\setlength{\belowrulesep}{0pt}
\begin{tabularx}{\textwidth}{p{0pt}@{\hspace{2.5pt}}
>{\raggedright\arraybackslash}p{120pt}
>{\raggedright\arraybackslash}X}
\toprule
\rule{0pt}{2.6ex}& \textbf{Aspect} & \textbf{Metrics} \\[2pt]
\midrule
\simgrp{cSem}{Semantic}
\simrow{cSem} Global semantic similarity & \simcc{cSem}Embedding Similarity \\
\simrow{cSem} Sentence-level correspondence & \simcc{cSem}InfoMatch (Sent) \\
\simrow{cSem} Lexical overlap & \simcc{cSem}BLEU, ROUGE-L \\
\simgap
\simgrp{cLex}{Lexical}
\simrow{cLex} Lexical characteristics & \simcc{cLex}Lexical Diversity (MTLD), Word Frequency (Zipf) \\
\simgap
\simgrp{cSyn}{Syntactic}
\simrow{cSyn} Grammatical profiles & \simcc{cSyn}POS Distribution, Dependency Distribution, Function Word Profile \& Density \\
\simgap
\simgrp{cStr}{Structural}
\simrow{cStr} Response organization & \simcc{cStr}Response Length (Chars), Paragraph Density \\
\simrow{cStr} Formatting conventions & \simcc{cStr}Header Use, List Use \& Density \\
\simgap
\simgrp{cSty}{Stylistic \& Presentational}
\simrow{cSty} Linguistic and affective profiles & \simcc{cSty}Readability, Concreteness, Emotion Profile \& Intensity, Punctuation Profile \& Density \\
\simrow{cSty} Interactional conventions & \simcc{cSty}Opening Template, Closing Template \\
\simrow{cSty} Visual emphasis & \simcc{cSty}Markdown Emphasis Use \& Density \\
\bottomrule
\end{tabularx}
\end{NoHyper}
\caption{Dimensions and aspects of textual similarity, with their corresponding metrics. Detailed metric definitions and computation methods are provided in Table~\ref{tab:framework_metrics} and Appendix~\ref{app:metric_implementation}.}
\label{tab:similarity_aspects}
\end{table*}

\section{Related Work}
\label{sec:related_work}

\paragraph{Characterizing and Measuring LLM Homogeneity.}
Prior work has documented substantial homogeneity in LLM generation, often quantified through embedding-based similarity \citep{jiang2025artificial, wenger2025we, patel2026llms}. Using real-world open-ended queries, \citet{jiang2025artificial} identify an \emph{Artificial Hivemind} effect, comprising intra-model repetition and, more critically, inter-model homogeneity, bringing output homogeneity to the forefront as a concern for LLM diversity and creativity. Beyond embedding-based similarity, \citet{guo2025benchmarking} systematically characterize linguistic diversity across semantic, lexical, and syntactic dimensions, while \citet{zhang2025noveltybench} and \citet{jain2025task} emphasize user-centered functional diversity. Extending the analysis across model generations, \citet{patel2026llms} report decreasing cross-model embedding distances from 2023 to 2026, while \citet{gude2026more} find reduced lexical and constructional diversity in newer instruction-tuned models. Building on these studies, we examine multidimensional textual similarity in relation to human judgments and trace its evolution both within and across models.

\paragraph{Effects of Post-Training on Response Diversity.}
Several studies investigate how post-training affects output diversity, often in the context of the alignment tax \citep{lin2024mitigating, padmakumar2024does, kirk2024understanding, murthy2025one}. While preference optimization has been associated with reduced output diversity, \citet{shypula2025evaluating} show that reductions in raw diversity do not necessarily imply lower effective diversity among high-quality responses. More recently, \citet{karouzos2026does} find that diversity collapse occurs at different stages depending on training data composition and post-training pipelines. These studies primarily examine within-model diversity, and we complement them by tracing cross-model differences across corresponding post-training stages.

\section{Operationalizing LLM Response Similarity Along Fine-Grained Dimensions}

In this section, we organize textual similarity into five dimensions and ten fine-grained aspects, then introduce 25 interpretable metrics to operationalize these aspects and measure similarity.

\subsection{Dimensions of Textual Similarity}
\label{sec:similarity_aspects}

Systemic Functional Linguistics (SFL) distinguishes what texts communicate from how meaning is expressed through vocabulary, grammar, and written form \citep{halliday2014introduction}. It also considers, through three metafunctions, how language conveys ideas, establishes social relationships, and organizes information. Together, these complementary perspectives highlight that texts can share broadly similar meanings while differing in their linguistic forms and organization.

Building on these distinctions, we consider five complementary dimensions of textual similarity. \textbf{Semantic} similarity encompasses global semantic similarity, sentence-level correspondence, and lexical overlap. \textbf{Lexical} similarity instead concerns vocabulary-use characteristics, including lexical diversity and word frequency, while \textbf{Syntactic} similarity concerns grammatical profiles. \textbf{Structural} similarity captures response organization and formatting conventions. Finally, \textbf{Stylistic and Presentational} similarity covers linguistic and affective profiles, interactional conventions, and visual emphasis. Together, these linguistically motivated dimensions provide broad, though not exhaustive, coverage of textual similarity.

\subsection{Measuring Textual Similarity}
We measure these dimensions with 25 interpretable metrics, grouped into ten aspects (Table~\ref{tab:similarity_aspects}).

Within the Semantic dimension, whole-response embedding similarity measures global semantic similarity, InfoMatch measures sentence-level information correspondence through bidirectional matching, and BLEU and ROUGE-L measure lexical overlap. We place BLEU and ROUGE-L under Semantic rather than Lexical because they measure content overlap between two responses, whereas Lexical metrics describe vocabulary properties of each response on its own.

Distinct from lexical overlap, lexical characteristics describe response-level properties of vocabulary use. Specifically, lexical diversity (MTLD) measures vocabulary richness, while mean Zipf frequency captures how commonly the words in a response are used. The remaining dimensions characterize grammatical patterns, response organization, and stylistic and presentational properties.

Structural characteristics encompass response organization, measured by response length and paragraph density, and formatting conventions, measured by header and list usage. Stylistic and presentational characteristics cover readability, concreteness, emotional expression, and punctuation patterns. We further distinguish interactional conventions, captured by opening and closing templates, from visual emphasis, captured by Markdown emphasis use and density. Together, these metrics characterize both broad linguistic patterns and specific response conventions that may vary across models. More details about how metrics are captured and computed can be found in Appendix~\ref{app:metric_implementation}.

\section{What Makes Two Responses Similar?}
\label{sec:human_embedding_check}
Before applying these metrics to study model homogeneity, we first examine which textual properties are associated with human perceptions of response similarity (\S\ref{sec:human_similarity}), then assess how well embedding similarity reflects these properties (\S\ref{sec:compared_with_embedding}).

\subsection{How Do Humans Perceive Similarity}
\label{sec:human_similarity}

\subsubsection{Experimental settings}

We conduct a two-round annotation study that separates overall similarity from meaning similarity and form similarity. We sample 90 response pairs from \textsc{Infinity-Chat}~\citep{jiang2025artificial},\footnote{\label{fn:infinity}We use the pairwise human-annotation subset of \textsc{Infinity-Chat} (500 response pairs; 50 prompts $\times$ 10 pairs): \url{https://huggingface.co/datasets/liweijiang/infinite-chats-human-pairwise}.} a WildChat-derived dataset of open-ended human--AI interactions~\citep{zhao2024wildchat}. We stratify pairs by \texttt{Qwen3-Embedding-8B}~\citep{qwen3embedding} cosine similarity and edit a subset to increase form variation while preserving semantic content.

\begin{figure}[t]
    \centering
    \includegraphics[
        width=0.7\columnwidth,
        trim=0 5 0 8,
        clip
    ]{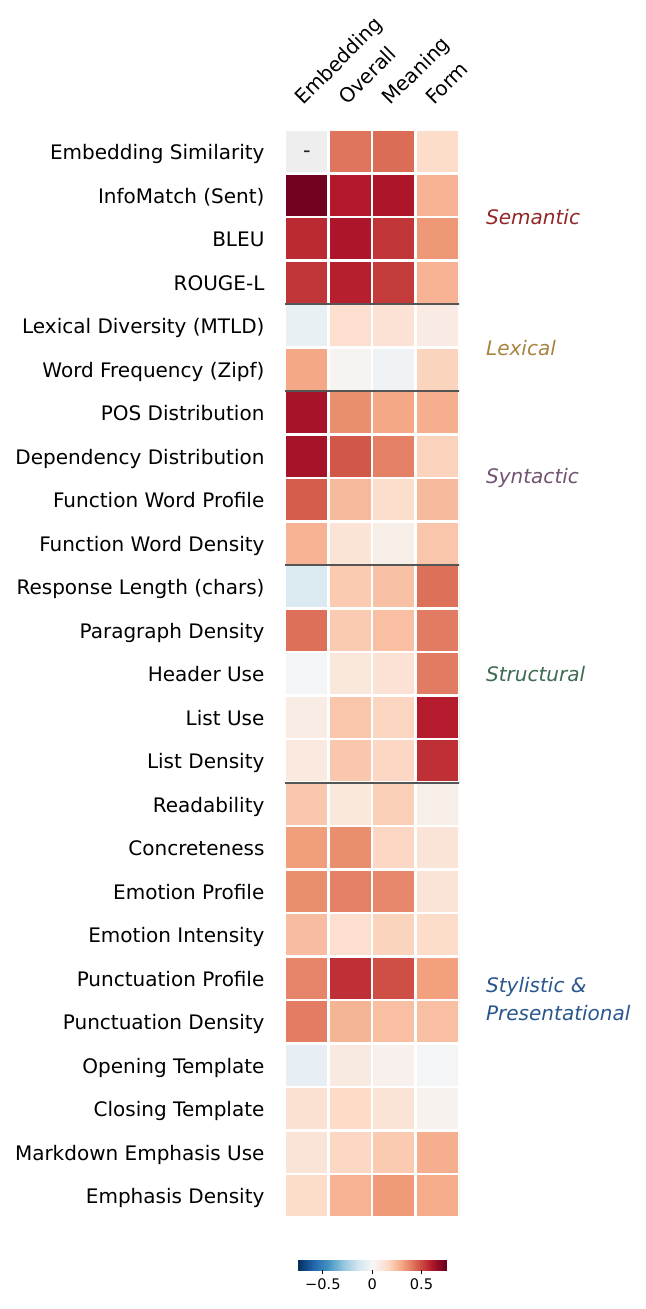} 
    \caption{Spearman correlations of similarity metrics with embedding similarity (first column; $n = 500$ pairs) and with human judgments of overall, meaning, and form similarity (remaining columns; $n = 90$ pairs; ratings averaged across three annotators). Exact values are reported in Table~\ref{tab:human_metric_corr} (human judgments) and Table~\ref{tab:embedding_robustness} (embedding similarity).}
    \label{fig:metric_correlations}
    \vspace{-0.5em}
\end{figure}

Eight annotators participate in the study. In the first round, annotators assess the \emph{overall similarity} of each response pair on a five-point scale, ranging from 1 (\emph{as different as two answers to this question could be}) to 5 (\emph{nearly identical}). In the second round, annotators evaluate the same response pairs separately in terms of \emph{meaning similarity} (what the responses say) and \emph{form similarity} (how the responses are written).
In each round, every response pair is independently evaluated by exactly three annotators. Each annotator evaluates 33 or 34 of the 90 main pairs, yielding 270 main annotations per round. Further details are provided in Appendix~\ref{app:human_annotation}.

\subsubsection{Human similarity analysis}

Human similarity judgments depend primarily on meaning, but form contributes independently.
Using partial Spearman correlations, we examine how meaning and form judgments contribute to overall similarity. Meaning dominates perceived similarity: it remains strongly associated with overall ratings after controlling for form ($\rho = 0.714$, $p < 0.001$). Nevertheless, form retains a significant association after controlling for meaning ($\rho = 0.326$, $p < 0.01$), indicating how responses are written matters beyond what they say.

Meanwhile, human meaning similarity aligns with specific informational correspondence and linguistic feature distributions. Embedding Similarity correlates moderately with human meaning judgments ($\rho = 0.417$), indicating that it does capture some aspects of perceived semantic similarity. However, InfoMatch, which uses the same embedding model but performs sentence-level matching, shows a numerically higher correlation ($\rho=0.606$), as do the lexical-overlap measures BLEU ($\rho=0.538$) and ROUGE-L ($\rho=0.525$). 

Beyond semantic measures, metrics capturing grammatical profiles, including POS and Dependency Distributions, also correlate with human meaning judgments, suggesting that grammatical patterns reflect aspects of meaning. Notably, Emotion and Punctuation Profiles further show higher observed correlations than their scalar counterparts, suggesting that distributional patterns carry more information about perceived meaning similarity than aggregate intensity or density alone.

Unlike meaning similarity, form judgments show their highest observed correlations with structural characteristics, including the adoption and intensity of formatting features and response-level organization. The strongest correlates are List Use ($\rho = 0.590$), List Density ($\rho = 0.548$), Response Length ($\rho = 0.412$), Header Use ($\rho = 0.389$), and Paragraph Density ($\rho = 0.388$).

\subsection{What Embeddings Capture and What They Miss}
\label{sec:compared_with_embedding}

\subsubsection{Experimental settings}
We use all 500 \textsc{Infinity-Chat} response pairs to examine which textual properties embeddings capture, with \texttt{Qwen3-Embedding-8B}~\citep{qwen3embedding} as the primary embedding model. Robustness checks with \texttt{Jina}~\citep{akram2026jina} and \texttt{QZhou-Embedding}~\citep{yu2025qzhouembeddingtechnicalreport} yield similar results (Appendix~\ref{app:embedding_robustness}).

\subsubsection{Relationships between metrics and embeddings}

From Figure~\ref{fig:metric_correlations} and Table~\ref{tab:embedding_robustness}, we can see embeddings primarily reflect semantic similarity, while also capturing some syntactic properties. Semantic metrics show the strongest correlations with embedding similarity, including InfoMatch, ROUGE-L, and BLEU ($\rho=0.63$--$0.65$). Syntactic metrics also show substantial associations, particularly POS and Dependency Distributions ($\rho=0.53$--$0.54$), followed by Function Word Profile ($\rho=0.39$). 

Notably, embeddings underrepresent features central to human form judgments: organization and formatting cues that strongly predict human ratings correlate only weakly with embedding similarity, including List Use ($\rho=0.139$), List Density ($\rho=0.177$), Response Length ($\rho=0.075$), and Header Use ($\rho=-0.020$).

These findings reveal a gap between human perception and embedding-based similarity: Humans consider information and response organization, while embeddings capture only part of the meaning and miss several structural and presentational features. Thus, high embedding similarity does not always mean humans see two responses as similar.

\section{What Do Current Models Share, and Where Do They Differ?}
\label{sec:current}

Section~\ref{sec:human_embedding_check} shows that human-perceived response similarity extends beyond what whole-response embeddings capture, particularly in textual form. With this distinction in mind, we revisit homogeneity among contemporary LLMs, examining which textual characteristics they share and where model-specific differences remain.

\subsection{Experimental Settings}

\paragraph{Models and prompts.} We sample 300 questions from \textsc{Infinity-Chat} (the full open-ended query set (26K)), balanced across six open-ended question categories. To avoid mechanically inducing similarity in the properties we aim to measure, we remove explicit stylistic and formatting constraints from the prompts, such as ``write in bullet points'' or ``in Hemingway's style.''

We evaluate 18 models. Fifteen are selected from five families—Gemma, Qwen, OLMo, GLM, and Granite—with regular releases from 2024 to 2026. We select one model per family for each year (also used in Section~\ref{sec:temporal}), prioritizing comparable model sizes. We additionally include three models outside this family-based temporal comparison: DeepSeek-V4-Flash~\citep{deepseekai2026deepseekv4}, Gemini-3.1-Flash-Lite~\citep{googledeepmind2026gemini31flashlite}, and GPT-5.4-Nano~\citep{openai2026gpt54mininano}.

\definecolor{meaningbg}{HTML}{DCECF8}
\definecolor{formbg}{HTML}{F8E9D7}

\begin{table}[!t]
\centering
\footnotesize
\setlength{\tabcolsep}{1.6pt}
\renewcommand{\arraystretch}{1.05}
\begin{tabular*}{\columnwidth}{@{\hspace{\tabcolsep}\extracolsep{\fill}}l>{\color{black!55}}r@{\hspace{6pt}}>{\color{black!55}}r@{\hspace{6pt}}>{\color{black!55}}r@{\hspace{6pt}}l}
\toprule
 & \multicolumn{3}{c}{\textbf{Distance}} & \\
\cmidrule(lr){2-4}
\textbf{Metric} & \hd{Intra} & \hd{Inter} & \hd{Rand} & \multicolumn{1}{l}{\textbf{\textit{$ N$}}} \\
\midrule
\multicolumn{5}{l}{\textit{Weak separation ($N>0.7$)}} \\
\cellcolor{meaningbg}Embedding Similarity & .101 & .176 & .895 & \nv{.905} \\
\cellcolor{meaningbg}InfoMatch (Sent)     & .263 & .321 & .559 & \nv{.805} \\
Emotion Profile                           & .062 & .115 & .325 & \nv{.797} \\
Concreteness                              & .045 & .084 & .218 & \nv{.776} \\
Emotion Intensity                         & .086 & .132 & .267 & \nv{.748} \\
POS Distribution                          & .016 & .034 & .077 & \nv{.715} \\
Function Word Profile                     & .018 & .036 & .079 & \nv{.704} \\
\addlinespace[4pt]
\multicolumn{5}{l}{\textit{Moderate separation ($0.4\le N\le0.7$)}} \\
\cellcolor{formbg}List Use                & .072 & .161 & .368 & \nv{.700} \\
Dependency Distribution                   & .038 & .064 & .122 & \nv{.683} \\
Function Word Density                     & 2.715 & 4.771 & 8.702 & \nv{.657} \\
\cellcolor{formbg}List Density            & .995 & 1.766 & 3.095 & \nv{.632} \\
Word Frequency (Zipf)                     & .089 & .164 & .292 & \nv{.631} \\
Readability                               & 1.092 & 2.329 & 3.987 & \nv{.573} \\
\cellcolor{formbg}Paragraph Density       & .707 & 1.288 & 1.963 & \nv{.538} \\
\cellcolor{meaningbg}Punctuation Profile  & .051 & .126 & .202 & \nv{.507} \\
\cellcolor{meaningbg}ROUGE-L              & .736 & .822 & .910 & \nv{.505} \\
\cellcolor{formbg}Response Length (Chars) & 486   & 1511  & 2296  & \nv{.433} \\
Emphasis Density                          & 1.102 & 2.366 & 3.284 & \nv{.421} \\
Punctuation Density                       & 2.961 & 6.197 & 8.429 & \nv{.408} \\
\addlinespace[4pt]
\multicolumn{5}{l}{\textit{Strong separation ($N<0.4$)}} \\
Markdown Emphasis Use                     & .073 & .235 & .327 & \nv{.364} \\
\cellcolor{meaningbg}BLEU                 & .837 & .939 & .992 & \nv{.341} \\
Lexical Diversity (MTLD)                  & 21.57 & 38.18 & 46.46 & \nv{.332} \\
\cellcolor{formbg}Header Use              & .080 & .404 & .507 & \nv{.242} \\
Opening Template                          & .068 & .324 & .357 & \nv{.115} \\
Closing Template                          & .100 & .304 & .327 & \nv{.101} \\
\bottomrule
\end{tabular*}
\caption{Intra-model, inter-model, and random-pair distances across 25 textual metrics for 18 models (153 unordered model pairs). All values are expressed as distances, with semantic similarities $S$ converted to $1-S$. $N$ (Eq.~\ref{eq:nis}) denotes normalized inter-model similarity. Lower $N$ indicates greater cross-model separation. Blue and orange shading mark the five metrics with the highest observed correlations with human meaning and form similarity judgments, respectively (\S~\ref{sec:human_similarity}). Bars visualize $N$ on a 0--1 scale.}
\label{tab:current_models}
\vspace{-0.5em}
\end{table}

Following \citet{jiang2025artificial}, we sample responses using temperature $1.0$ and top-$p$ $0.9$ to examine response diversity. As incoherent generations can bias homogeneity measurements~\citep{shypula2025evaluating}, we filter out responses that are irrelevant, incomplete, or contain leaked metadata or reasoning traces. After filtering, 279 prompts remain, with at least 30 valid responses per model for each prompt.\footnote{We jointly filter responses for this section and Section~\ref{sec:temporal}, retaining the same 279 prompts and 30 valid responses per model per prompt for both analyses.} 
Details of models, prompts and filtering process are provided in Appendix~\ref{app:models_response}.

\paragraph{Intra-/Inter-Model Distances and Normalized Inter-Model Similarity.}
We orient all metrics as distances, so larger values indicate greater differences between responses. For each prompt, we compute the average distance between responses from the same model ($D_{\mathrm{intra}}$) and from different models ($D_{\mathrm{inter}}$), then average over all prompts. We also compute a random baseline ($D_{\mathrm{random}}$) from 10,000 response pairs sampled across all prompts and models, which serves as an empirical upper bound for each metric.

\begin{figure*}[t]
    \centering
    \includegraphics[width=\textwidth]{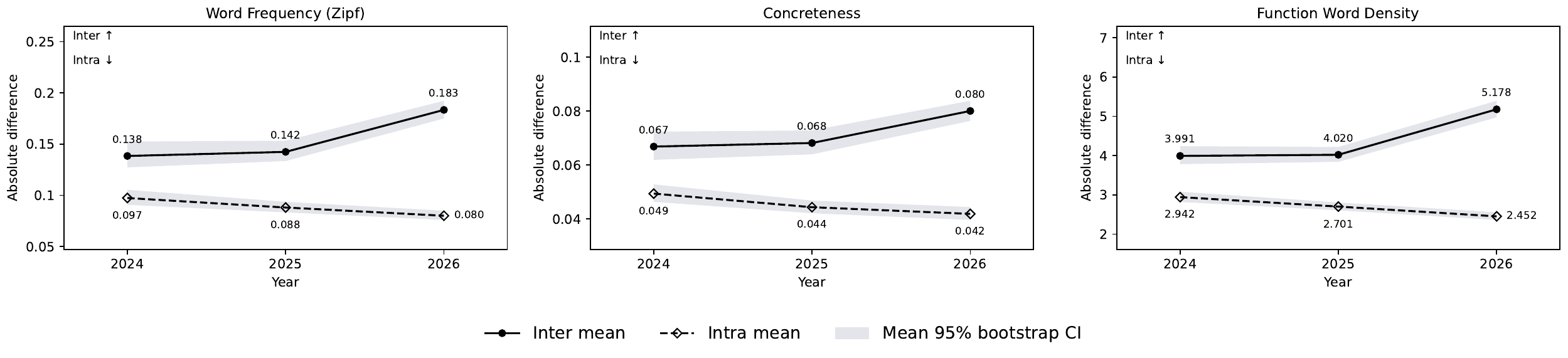}
    \caption{Inter- and intra-model distances from 2024 to 2026 for three representative metrics. On all three, models become more like themselves (decreasing intra-model distance) but less like one another (increasing inter-model distance). See Figure~\ref{fig:temporal_trends} for all metrics.}
\label{fig:temporal_inter_intra}
\end{figure*}

Motivated by the observation that responses to the same prompt already share language, topic, length, and register \citep{schaeffer2026strong}, we compare $D_{\mathrm{inter}}$ with $D_{\mathrm{intra}}$, which are both computed within prompts and thus equally affected by this shared similarity. 

To quantify this comparison, we define the normalized similarity score $N$ as

\begin{equation}
N =
\frac{D_{\mathrm{random}} - D_{\mathrm{inter}}}
     {D_{\mathrm{random}} - D_{\mathrm{intra}}}.
\label{eq:nis}
\end{equation}

$N$ measures how close the inter-model distance is to the intra-model distance relative to the random-pair reference. A value of $N=1$ indicates that inter-model and intra-model distances are equal, while $N=0$ indicates that inter-model distance equals the random-pair distance. 

\subsection{Results}

\paragraph{Models say similar things in different words.}

As shown in Table~\ref{tab:current_models}, Embedding Similarity and InfoMatch (Sent) achieve the highest $N$ scores (0.905 and 0.805), indicating that inter-model distances remain close to intra-model distances relative to the random-pair reference. In contrast, BLEU and ROUGE-L, which capture lexical overlap, exhibit substantially lower $N$ scores (0.341 and 0.505), revealing greater cross-model separation in wording. This distinction between semantic similarity and lexical differences is broadly consistent across question types (Figure~\ref{fig:question_type}).

\paragraph{Models share broad linguistic and affective profiles.} Beyond semantic correspondence, models exhibit substantial similarity in several grammatical, linguistic, and affective characteristics, including POS and dependency distributions, emotion profiles, and concreteness. Presentational conventions show more varied patterns: List Use exhibits moderate cross-model separation ($N=0.700$), whereas Markdown Emphasis Use shows much stronger separation ($N=0.364$).

\paragraph{Models retain distinct lexical characteristics, structural patterns, and interactional conventions.} Despite these shared characteristics, substantial cross-model differences remain in other aspects. For example, Lexical Diversity (MTLD; $N=0.332$) exhibits strong separation, while Response Length ($N=0.433$) shows moderate separation. Interactional conventions are even more distinctive, with Opening and Closing Templates exhibiting the lowest $N$ scores among all metrics (0.115 and 0.101), indicating that their inter-model distances are much closer to the random-pair reference than to intra-model distances.

\section{How Does LLM Homogeneity Evolve?}
\label{sec:evolve}

Section~\ref{sec:current} identifies three patterns of cross-model similarity: shared semantic content, common linguistic profiles, and distinctive response conventions. We next trace how these patterns evolve across generations from 2024 to 2026 (\S\ref{sec:temporal}) and across post-training stages (\S\ref{sec:stage}). Because a change in $N$ can reflect either weaker inter-model homogeneity or stronger intra-model repetition, we analyze $D_{\mathrm{inter}}$ and $D_{\mathrm{intra}}$ separately throughout.

\subsection{Across Model Generations}
\label{sec:temporal}

We analyze fifteen open-weight models from the Gemma, Qwen, OLMo, GLM, and Granite families, with one release per family from each year between 2024 and 2026, using the same prompts and sampling configuration as in Section~\ref{sec:current}.

\definecolor{incgreen}{RGB}{26,127,55}
\definecolor{decred}{RGB}{196,40,40}
\newcommand{\inc}[1]{\textcolor{incgreen}{$\blacktriangle$\,+#1\%}}
\newcommand{\dec}[1]{\textcolor{decred}{$\blacktriangledown$\,$-$#1\%}}
\newcommand{\nsrow}[1]{\textcolor{gray}{#1}}

\textbf{Newer models become more self-consistent, yet more distinct from one another along many metrics.}
From 2024 to 2026, intra-model distances change significantly for 22 of the 25 metrics, decreasing in 17 (77\%).\footnote{a 2024--2026 change is considered significant if its 95\% CI excludes zero, based on 10,000 prompt-level bootstrap resamples (prompts resampled jointly across models and years).}This trend spans all five textual dimensions, extending the single-snapshot, embedding-based evidence of intra-model repetition reported by \citet{jiang2025artificial}.

In contrast, inter-model distances change significantly for 21 metrics, increasing in 14 of them (67\%). Notably, several metrics exhibit decreasing intra-model distances alongside increasing inter-model distances (Figure~\ref{fig:temporal_inter_intra}). For example, Word Frequency (intra: 0.097$\to$0.080; inter: 0.138$\to$0.183) and Function Word Density (intra: 2.942$\to$2.452; inter: 3.991$\to$5.178) move in opposite directions. 

These findings reveal that models can become more like themselves while becoming less like one another. The two facets of the \emph{Artificial Hivemind} thus evolve differently: intra-model repetition intensifies, whereas inter-model homogeneity, which \citet{jiang2025artificial} regard as the more critical concern, weakens along many textual dimensions.

\begin{table}[t]
\centering
\small
\setlength{\tabcolsep}{8pt}
\renewcommand{\arraystretch}{1.08}
\begin{tabular}{@{}lr@{}}
\toprule
\textbf{Metric} &
\textbf{$\Delta D_{\mathrm{inter}}$ (2024$\to$2026)} \\
\midrule
\multicolumn{2}{@{}l}{\textit{Human-Salient Meaning Features}} \\
Embedding Similarity                      & \dec{10.0} \\
InfoMatch                                 & \inc{9.6} \\
\nsrow{Punctuation Profile$^{\dagger}$}
                                          & \nsrow{$\blacktriangle$\,+7.1\%} \\
ROUGE-L                                   & \inc{3.2} \\
BLEU                                      & \inc{1.2} \\
\midrule
\multicolumn{2}{@{}l}{\textit{Human-Salient Form Features}} \\
List Use                                  & \dec{39.8} \\
Paragraph Density                         & \dec{22.3} \\
List Density                              & \dec{21.1} \\
Response Length                           & \inc{31.5} \\
\nsrow{Header Use$^{\dagger}$}
                                          & \nsrow{$\blacktriangle$\,+8.0\%} \\
\midrule
\multicolumn{2}{@{}l}{\textit{Lexical Characteristics and Interactional Conventions}} \\
Closing Template                          & \inc{169.0} \\
Opening Template                          & \inc{85.1} \\
Lexical Diversity (MTLD)                  & \inc{78.0} \\
Word Frequency (Zipf)                     & \inc{32.5} \\
\bottomrule
\end{tabular}
\caption{
Changes in inter-model distance from 2024 to 2026.
Human-salient meaning and form features denote the five metrics with the highest observed correlations with human meaning and form judgments (\S~\ref{sec:human_similarity}). $\Delta D_{\mathrm{inter}}$ is the relative change from 2024: \textcolor{incgreen}{$\blacktriangle$} means models diverge and \textcolor{decred}{$\blacktriangledown$} means they converge. Grey entries ($^{\dagger}$) have 95\% prompt-level bootstrap CIs (10{,}000 resamples) that include zero.}
\label{tab:temporal_grouped}
\vspace{-10pt}
\end{table}

\begin{figure*}[!t]
    \centering
    \includegraphics[width=\textwidth]{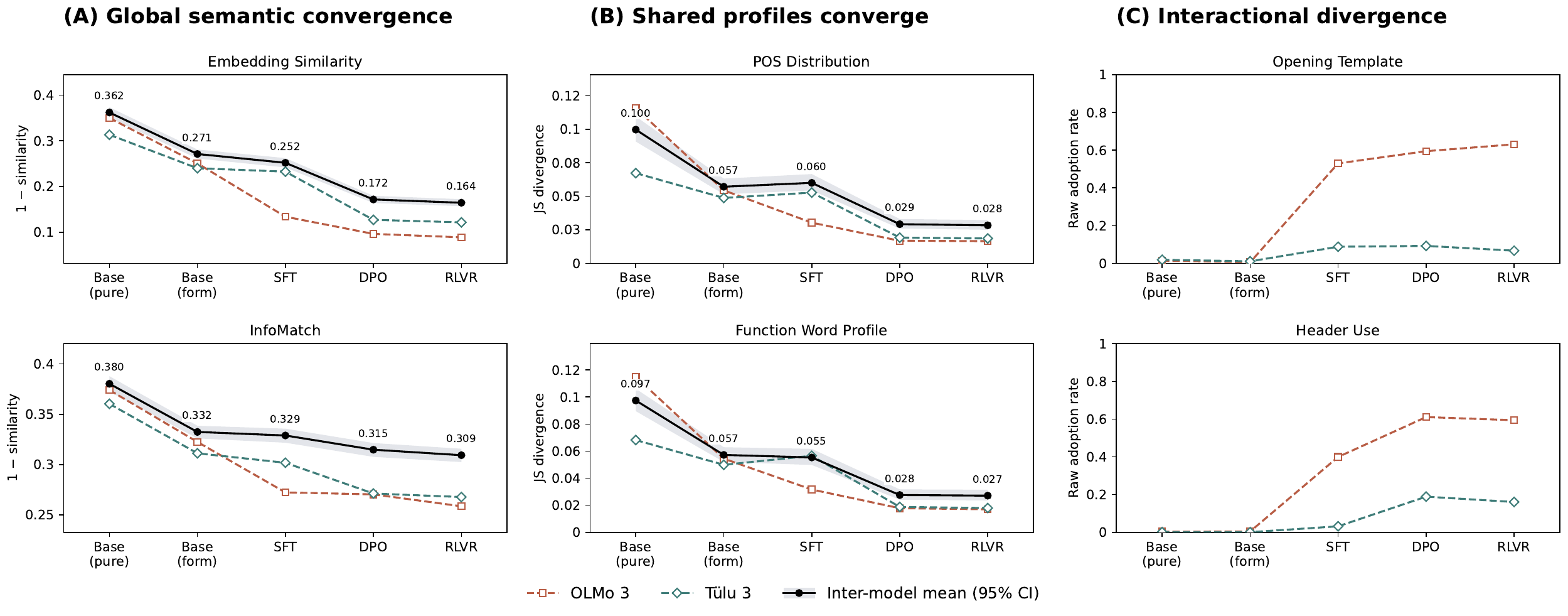}
    \caption{Post-training narrows semantic and grammatical distances while interactional and formatting conventions diverge. Training-stage trends for six representative metrics in OLMo~3 and T\"ulu~3. Dashed lines show adoption rates for Opening Template/Header Use and intra-model distances otherwise; black lines show mean inter-model distances with 95\% bootstrap CIs shaded. Lower distances indicate greater similarity. Base (pure/form) uses three-shot demonstrations without/with formatting markers. See Figure~\ref{fig:stage_full} for all metrics.}
    \label{fig:stage_representative}
\vspace{-5pt}
\end{figure*}

\textbf{Global semantic convergence masks growing inter-model differences.}
Consistent with prior work, embedding similarity suggests strong cross-model homogeneity both synchronically \citep{wenger2025we, jiang2025artificial} and diachronically \citep{patel2026llms}: it yields the highest normalized similarity among all 25 metrics ($N = 0.905$; \S\ref{sec:current}), and inter-model embedding distance decreases by 10.0\% from 2024 to 2026. However, this convergence is not reflected in finer-grained semantic measures: Sentence-level InfoMatch distance increases by 9.6\%, while BLEU and ROUGE-L distances also increase. These contrasting trends suggest that embedding similarity alone may increasingly overstate inter-model homogeneity.

\textbf{Human-salient form features partly converge, while model-specific conventions diverge.} Among the five metrics most strongly correlated with human form judgments (\S\ref{sec:human_similarity}), three show decreasing inter-model distances: List Use ($-39.8\%$), Paragraph Density ($-22.3\%$), and List Density ($-21.1\%$). However, the largest cross-generational divergences occur in lexical characteristics and interactional conventions, which correlate weakly with human form judgments (Table~\ref{tab:temporal_grouped}). For example, inter-model distances increase by 169.0\% for Closing Template and 78.0\% for Lexical Diversity. Together, models increasingly share how they organize responses, while becoming more distinct in their lexical and interactional conventions.

\subsection{Across Post-Training Stages}
\label{sec:stage}
To understand how the three patterns of cross-model similarity identified in Section~\ref{sec:current} evolve during post-training, we compare corresponding checkpoints from the OLMo~3 and T\"ulu~3 pipelines.\footnote{Because base models are elicited with few-shot demonstrations, we treat them as a reference rather than a directly comparable stage, and focus on changes among the post-trained checkpoints (SFT, DPO, and RLVR), which share the same generation setup.}
Experimental details are provided in Appendix~\ref{app:settings_for_stage}.

\paragraph{Semantic convergence continues beyond supervised fine-tuning.} As observed in \S~\ref{sec:current}, current models share substantial semantic content, while Section~\ref{sec:temporal} shows that global semantic similarity has increased across model generations. Tracing two post-training pipelines, we find that semantic convergence continues during preference optimization. Inter-model embedding distance decreases from 0.252 at SFT to 0.172 at DPO, with a smaller further reduction to 0.164 at RLVR. Sentence-level InfoMatch, BLEU, and ROUGE-L show consistent increases in cross-model similarity. Unlike the mixed generational trends, preference optimization reduces inter-model distances across all three semantic aspects: global semantic similarity, sentence-level correspondence, and lexical overlap. 

\paragraph{SFT introduces distinctive response patterns, while later stages reshape them unevenly.}
Sections~\ref{sec:human_embedding_check} and~\ref{sec:current} identify response organization and formatting conventions as important to human-perceived form similarity, alongside model-specific lexical characteristics and interactional conventions. Our stage-wise analysis shows that many of these differences become pronounced during SFT. Later stages, however, reshape these differences in different directions: by RLVR, Header and Opening/Closing Template mismatches increase further, while List Density differences decrease, and MTLD distance increases at DPO before declining at RLVR. Unlike a uniform diversity-collapse account of post-training, these findings reveal different trajectories across textual dimensions: preference optimization continues to increase semantic similarity while selectively amplifying or reducing model-specific differences in textual form.

\section{Conclusion}

We revisit LLM homogeneity through a multidimensional lens, using 25 metrics across five textual dimensions to separate what models say from how they say it. Our human study shows that perceived similarity depends on both, whereas whole-response embeddings capture content only partially and form hardly at all.

Viewed through this lens, LLM homogeneity emerges as a multidimensional and evolving phenomenon, where semantic convergence, structural alignment, and model-specific differentiation coexist. These patterns emerge at three levels of comparison. Across current models, LLMs say similar things in different words: they share global semantics and broad linguistic characteristics but retain distinct lexical characteristics, structural patterns, and interactional conventions. Across generations from 2024 to 2026, models have become more like themselves but not more like one another. Intra-model repetition increases for most metrics, while inter-model differences widen in fine-grained content, wording, and interactional conventions, even as global embedding similarity rises and several human-salient organizational features converge. Across post-training stages, these conventions emerge during SFT and are selectively reshaped, rather than uniformly collapsed, by preference optimization.

These findings caution against relying on embedding similarity alone, which reduces homogeneity to a single convergent trend and obscures metrics that move in other directions. More broadly, evaluations of LLM homogeneity should distinguish content from form, and intra-model repetition from inter-model similarity, when tracing how responses evolve across generations and post-training stages.

\newpage
\section*{Limitations}

\paragraph{Language, prompts, and decoding.}
We focus on English open-ended prompts from \textsc{Infinity-Chat}, since open-ended generation is where homogeneity concerns are most pressing, and several of our metrics rely on linguistic resources well established mainly for English. We adopt the decoding configuration of \citet{jiang2025artificial} (temperature 1.0, top-p 0.9) to ensure comparability with prior work. Because temperature affects within-model variability, absolute distances may shift under other decoding settings. Extending the framework to other languages, task types, and decoding strategies is left for future work.

\paragraph{Model coverage and confounds.}
Our cross-model analysis spans 18 models, yielding inter-model distances over 153 model pairs. For the generational analysis, few families offer open-weight releases in every year from 2024 to 2026, so model size and architecture are not held constant across releases. We therefore select the closest available sizes and evaluate hybrid-reasoning models in non-thinking mode for comparability. The resulting trends are descriptive and cannot be attributed to any single factor. We use lightweight API models because reliably estimating intra-model distances requires dozens of samples per prompt (at least 30 in our design), making larger proprietary systems costly to cover; such systems may exhibit different patterns.

\paragraph{Normalized similarity.}
Raw metric values lack an absolute scale for how similar two responses are. Even for embedding similarity, the same score can mean different things under different embedding models. We therefore use random response pairs as an empirical upper bound on distance, which requires no metric-specific assumptions and places heterogeneous metrics on a common scale. However, because random pairs pool responses across prompts and models, this reference is not a strict bound for prompt-insensitive metrics. We therefore treat $N$ as a relative ranking and report raw distances alongside it. The analyses in Section~\ref{sec:evolve} rely on raw distances rather than $N$.

\section*{Acknowledgments}
This research was supported by the Humanities and AI Virtual Institute (HAVI), funded by Schmidt Sciences, under the project Artificial Intelligence for Cultural and Historical Reasoning.

\bibliography{custom}

\clearpage
\appendix

\section{Metric Implementation Details}
\label{app:metric_implementation}
 
\subsection{Text Preprocessing and Metric-Specific Views}

Different metrics require different response representations. Whole-response embeddings and response length are computed on raw responses, whereas sentence-level correspondence and lexical overlap use content-focused representations that reduce the influence of markup and non-prose elements. Linguistic measurements are restricted to natural-language prose, while structural and presentational measurements preserve the relevant layout and formatting cues. We therefore build a separate response view for each group of metrics before computing features (Table~\ref{tab:text_views}). 
 
\begin{table*}[t]
\centering
\footnotesize
\renewcommand{\arraystretch}{1.2}
\begin{tabularx}{\textwidth}{@{}l
  >{\raggedright\arraybackslash}X
  >{\raggedright\arraybackslash}X
  >{\raggedright\arraybackslash}X@{}}
\toprule
\textbf{View} & \textbf{Processing} & \textbf{Metrics} & \textbf{Rationale} \\
\midrule
 
Raw response
& No preprocessing; the complete model output is retained
& Embedding Similarity; Response Length
& These metrics are explicitly defined over the response as originally produced \\
 
Semantic-content view
& Markdown syntax and URLs are removed; fenced code blocks are excluded; inline-code content and textual table-cell content are retained
& BLEU; ROUGE-L
& Preserves linguistic content while reducing overlap caused by markup or large code blocks \\
 
Sentence-semantic view
& Uses semantic-content blocks, additionally excludes tables, and segments each remaining block into sentences independently
& InfoMatch (Sent)
& Sentence-level correspondence requires natural sentence units; table rows can otherwise create artificial best matches \\
 
Prose view
& Markdown syntax, URLs, tables, fenced code blocks, and inline code are excluded; natural-language textual content is retained
& MTLD; Word Frequency; POS Distribution; Dependency Distribution; Function Word Profile/Density; Readability; Concreteness; Emotion Profile/Intensity; Punctuation Profile/Density
& Restricts linguistic and stylistic measurements to natural-language prose and avoids contamination from code or table syntax \\
 
Structural view
& Fenced code blocks are collapsed to single structural blocks and horizontal rules are removed; the remaining layout is preserved
& Paragraph Density
& Preserves document organization while preventing blank lines inside code blocks from creating artificial paragraphs \\
 
Markdown structure
& The original Markdown structure is parsed directly; code regions are excluded when identifying additional nonstandard bullet markers
& Header Use; List Use/Density; Markdown Emphasis Use/Density
& These metrics depend on explicit presentational structure that is removed from the prose representation \\
 
\bottomrule
\end{tabularx}
\caption{
Metric-specific response representations used during feature extraction.
Each view retains information relevant to its associated metrics while removing content that could introduce measurement artifacts.
}
\label{tab:text_views}
\end{table*}
 
\subsection{Metric Computation Details}
\label{sec:metric_computation}
 
Table~\ref{tab:framework_metrics} provides the high-level definitions of all metrics. Here we describe implementation choices that are important for reproducibility.

\paragraph{Shared lexical tokenization and density normalization.}
We use a common lexical-word definition across all word-normalized metrics. A lexical word is defined as a maximal sequence of Unicode letters, optionally containing an internal apostrophe. All density metrics (Function Word, Paragraph, List, Punctuation, and Emphasis Density) are computed per 100 lexical words in the prose view.
 
\paragraph{Sentence-level information matching.}
InfoMatch(sent) measures how well the sentences in one response can be matched to sentences in the other, regardless of order. Given sentences $S_A=\{a_1,\ldots,a_m\}$ and $S_B=\{b_1,\ldots,b_n\}$ with Qwen3-Embedding-8B representations $e(\cdot)$, we compute
\begin{align}
P &= \frac{1}{m}\sum_{i=1}^{m}\max_{j}\cos\bigl(e(a_i),e(b_j)\bigr), \\
R &= \frac{1}{n}\sum_{j=1}^{n}\max_{i}\cos\bigl(e(a_i),e(b_j)\bigr),
\end{align}
and take their harmonic mean, $\mathrm{InfoMatch} = 2PR/(P+R)$. This follows the matching idea of BERTScore~\cite{zhang2020bertscore}, applied to sentence embeddings instead of token embeddings.
 
\paragraph{Lexical overlap metrics.}
BLEU is computed using SacreBLEU sentence scoring with lowercasing, up-to-4-gram overlap, effective order, exponential smoothing, and the standard brevity penalty~\cite{papineni2002bleu}. Because BLEU is directional, we compute both $A\!\rightarrow\!B$ and $B\!\rightarrow\!A$ and use their arithmetic mean. ROUGE-L is computed as the F1 score induced by the longest common subsequence of the two semantic-content representations~\cite{lin2004rouge}.
 
\paragraph{Lexical statistics.}
Lexical Diversity is measured using MTLD~\cite{mccarthy2010mtld}. Because MTLD is unstable for very short samples, it is treated as undefined for responses containing fewer than 50 lexical words.
 
Word Frequency is the mean Zipf frequency of all lexical words according to \texttt{wordfreq}~\cite{robyn_speer_2022_7199437,vanheuven2014subtlex}. Words absent from the frequency resource receive a value of zero and remain in the average rather than being discarded.
 
\paragraph{Syntactic statistics.}
We parse each prose block independently with spaCy~\cite{honnibal2020spacy}
(\texttt{en\_core\_web\_sm} pipeline). POS Distribution is the normalized distribution over coarse part-of-speech tags among non-punctuation tokens, while Dependency Distribution is the corresponding distribution over dependency-relation labels.
 
The Function Word Profile contains seven categories: articles, prepositions, coordinating conjunctions, subordinating conjunctions, pronouns, auxiliaries, and particles. Articles are detected lexically, while the remaining categories are derived from spaCy POS annotations. 
 
\paragraph{Readability and concreteness.}
Readability is measured using Flesch--Kincaid grade level~\cite{kincaid1975flesch}. Each prose block is parsed independently when determining sentence boundaries.
 
Concreteness is estimated using a regression model trained from Brysbaert et al.'s word-concreteness ratings~\cite{brysbaert2014concreteness}, with FastText vectors used as word representations~\cite{bojanowski2017fasttext}. Predicted scores are constrained to the original 1--5 concreteness scale and averaged over lexical words for which a FastText representation is available.
 
\paragraph{Emotion.}
We apply the GoEmotions classifier~\cite{demszky2020goemotions} to the prose view. Long responses are split into chunks of up to 510 subwords, and chunk-level probabilities are averaged with weights proportional to chunk length. Emotion Profile is the normalized distribution over the 28 categories; Emotion Intensity is the highest probability among non-neutral categories.
 
\paragraph{Punctuation and emphasis measurements.}
Punctuation Profile is the distribution over 11 punctuation classes; punctuation used in numbers, such as commas in ``1,000'' or hyphens in ``2--3'', is not counted. For Emphasis Density, nested bold or italic markup is counted as one span.
 
\paragraph{Formatting and visual-emphasis indicators.}
Header Use, List Use, and Markdown Emphasis Use are binary indicators detected by parsing the original response with \texttt{markdown-it-py}. List Use and List Density also count bullet-marked lines the parser misses, detected with regular expressions. List Density and Emphasis Density are computed from the same parsed structures.

\paragraph{LLM-annotated interactional templates.}
Opening Template and Closing Template depend on the communicative role of the text and are therefore annotated using Qwen-3.8-27b \citep{qwen38} on the original response. The classifier returns two binary labels indicating whether the response begins with a formulaic interactional opener or ends with a formulaic closing or invitation for further interaction.

\begin{tcolorbox}[
colback=gray!5,
colframe=gray!50!black,
title=Prompt: Interactional Template Annotation,
breakable,
width=\linewidth
]
You will be given a model-generated response. Identify whether the response contains each of the following features.

1. \textbf{opening\_template}: The response begins with a phatic or meta-comment before addressing the question (e.g., ``That's a great question'', ``Sure!'', ``Certainly!'', ``Of course!'', ``Absolutely!''). Do not count responses that begin directly with substantive content.

2. \textbf{closing\_template}: The response ends with a formulaic offer or sign-off (e.g., ``Let me know if you have any questions'', ``I hope this helps'', ``Feel free to ask if you need clarification''). Do not count responses that end with substantive content.

Return only a JSON object with exactly these two Boolean fields:

\texttt{\{"opening\_template": boolean,}\\
\texttt{\phantom{\{}"closing\_template": boolean\}}
\end{tcolorbox}

\paragraph{Similarity Computation.} Our metrics produce four types of outputs: pairwise, scalar, distributional, and binary. Table~\ref{tab:pairwise_computation} summarizes how each is converted into pairwise similarity, where $x_i=m(r_i)$ denotes the measurement of response $r_i$. Native pairwise measures are symmetrized when necessary. All scores are oriented so that larger values indicate greater similarity, and are retained individually rather than aggregated into a single score.

\begin{table}[t]
\centering
\footnotesize
\setlength{\tabcolsep}{3.5pt}
\renewcommand{\arraystretch}{1.15}
\begin{tabular}{@{}lll@{}}
\toprule
\textbf{Output type} & \textbf{Example metric} & \textbf{Similarity} \\
\midrule
Pairwise       & Embedding cosine & Native score \\
Scalar         & Response Length  & $-|x_i-x_j|$ \\
Distributional & POS Distribution & $-\mathrm{JSD}(x_i,x_j)$ \\
Binary         & Header Use       & $\mathbb{I}[x_i=x_j]$ \\
\bottomrule
\end{tabular}
\caption{Pairwise similarity computation by metric output type, with representative examples.}
\label{tab:pairwise_computation}
\end{table}

\newcommand{\mdh}[1]{\par\vspace{2pt}\noindent{\bfseries #1}\par}                 %
\newcommand{\mdhh}[1]{\par\vspace{1pt}\noindent{\bfseries\itshape #1}\par}         %
\newcommand{\mdcode}[1]{\par\noindent\hspace*{1em}\texttt{#1}\par}    %
\newlist{mdlist}{itemize}{1}
\setlist[mdlist]{label=\textbullet, leftmargin=1.1em, itemsep=0pt, topsep=1pt, parsep=0pt}
\newcommand{\responsebox}[3]{%
  \fbox{\begin{minipage}[t]{\dimexpr#1\textwidth-2\fboxsep-2\fboxrule\relax}
    \footnotesize\setlength{\parskip}{2pt}%
    {\small\bfseries #2}\par\smallskip
    #3
  \end{minipage}}}

\section{Motivating Example}
\label{app:case-studies}

This section provides the full response pair shown in Figure~\ref{fig:motivating-example}, together with its human ratings and a summary of how responses differ in content.

\paragraph{Responses.}
Figure~\ref{fig:case-study-responses} shows both responses, with Markdown rendered as it would appear to a reader.

\begin{figure*}[!t]
\centering
\small\textbf{Prompt:} \textit{Give an example of a linear graph in graph theory.}
\par\smallskip
\responsebox{0.48}{Response 1}{%
In graph theory, a \textbf{linear graph} typically refers to a \textbf{path graph}, which is a simple type of graph where all vertices are arranged in a single straight line. Each vertex (except for the endpoints) is connected to exactly two other vertices, forming a linear sequence without any branches or cycles.

\mdh{Example of a Linear (Path) Graph}
Consider the \textbf{path graph \(P_4\)}, which consists of 4 vertices connected in a single line. Here's how it's structured:
\begin{mdlist}
  \item \textbf{Vertices:} \(V = \{v_1, v_2, v_3, v_4\}\)
  \item \textbf{Edges:} \(E = \{\{v_1, v_2\}, \{v_2, v_3\}, \{v_3, v_4\}\}\)
\end{mdlist}

\mdhh{Visualization:}
\mdcode{v1 \textrm{---} v2 \textrm{---} v3 \textrm{---} v4}

\mdhh{Properties of \(P_4\):}
\begin{mdlist}
  \item \textbf{Number of Vertices (n):} 4
  \item \textbf{Number of Edges (m):} 3
  \item \textbf{Degree Sequence:} Each interior vertex (\(v_2, v_3\)) has degree 2, while the endpoint vertices (\(v_1, v_4\)) have degree 1.
  \item \textbf{Path Length:} The length of the path is 3 (the number of edges).
  \item \textbf{Acyclic:} \(P_4\) does not contain any cycles.
  \item \textbf{Connected:} There is exactly one path between any two vertices.
\end{mdlist}

\mdh{General Definition}
A \textbf{path graph} \(P_n\) is a linear graph consisting of \(n\) vertices connected in a single straight path. Its general structure can be represented as:
\mdcode{v1 \textrm{---} v2 \textrm{---} v3 \textrm{---} ... \textrm{---} vn}
Where:
\begin{mdlist}
  \item Each \(v_i\) is a vertex.
  \item Each edge connects \(v_i\) to \(v_{i+1}\) for \(1 \leq i < n\).
\end{mdlist}

\mdh{Applications}
Path graphs are fundamental in graph theory and serve as building blocks for more complex structures. They are used in:
\begin{mdlist}
  \item \textbf{Network Design:} Modeling linear networks like pipelines or railway tracks.
  \item \textbf{Algorithm Design:} Serving as examples or test cases for graph algorithms.
  \item \textbf{Mathematical Proofs:} Providing simple cases for proving properties related to trees and connectivity.
\end{mdlist}

\mdh{Related Concepts}
\begin{mdlist}
  \item \textbf{Cycle Graphs:} Similar to path graphs but with an additional edge connecting the last vertex back to the first, forming a closed loop.
  \item \textbf{Star Graphs:} A type of tree graph with one central vertex connected to all other vertices, which are leaves.
  \item \textbf{Line Graphs:} A different concept where each vertex of the line graph represents an edge of the original graph, and adjacency is defined based on shared endpoints in the original graph.
\end{mdlist}

\mdh{Conclusion}
A linear (path) graph is one of the simplest forms of graphs in graph theory, characterized by its straightforward, straight-line structure. Understanding path graphs is essential as they form the basis for more intricate graph structures and are widely applicable in various fields such as computer science, biology, and engineering.
}\hfill
\responsebox{0.48}{Response 2}{%
In graph theory, a linear graph is a graph in which every vertex has degree 2. This means that each vertex is connected to exactly two edges.

Here's an example of a linear graph:

Suppose we have a graph with 5 vertices labeled A, B, C, D, and E. The edges of the graph are:
\begin{mdlist}
  \item A is connected to B
  \item B is connected to C
  \item C is connected to D
  \item D is connected to E
\end{mdlist}

This graph can be visualized as a straight line:

A -{}- B -{}- C -{}- D -{}- E

In this graph, each vertex has degree 2, meaning that each vertex is connected to exactly two edges. This is an example of a linear graph.

Note that a linear graph is also sometimes referred to as a ``path'' in graph theory.
}
\caption{Full response pair for the motivating example in Figure~\ref{fig:motivating-example}, with Markdown rendered.}
\label{fig:case-study-responses}
\end{figure*}

\paragraph{Human ratings.}
Table~\ref{tab:case-study-ratings} reports the original \textit{Infinity-Chat} ratings alongside the similarity ratings from our study.

\begin{table}[h]
\centering
\small
\begin{tabular}{lc}
\toprule
Rating & Score \\
\midrule
\multicolumn{2}{l}{\textit{Infinity-Chat (response quality)}} \\
Absolute, Response 1 (1--5) & 3.20 $\pm$ 1.26 \\
Absolute, Response 2 (1--5) & 3.92 $\pm$ 0.74 \\
Pairwise ($-2$ to $2$) & $-1.00$ \\
\midrule
\multicolumn{2}{l}{\textit{Our study (response similarity)}} \\
Overall (1--5) & 2 \; (3, 1, 2) \\
Meaning (1--5) & 3 \; (4, 3, 2) \\
Form (1--5) & 1 \; (1, 1, 1) \\
\midrule
Embedding similarity & 0.852 \\
\bottomrule
\end{tabular}
\caption{Human ratings and Qwen3-Embedding-8B similarity for the motivating example. Infinity-Chat ratings come from 25 annotators and assess response quality; absolute ratings are reported as mean $\pm$ standard deviation and the pairwise rating as the mean. Our study collects similarity ratings from three annotators, reported as the mean followed by individual ratings.}
\label{tab:case-study-ratings}
\end{table}

\paragraph{Content differences.}
Both responses illustrate a linear graph as a path and note the term ``path,'' but they differ in content in three ways. First, the definitions are incompatible: Response~1 states that endpoints have degree 1, whereas Response~2 requires every vertex to have degree 2, which describes a cycle and contradicts its own example, where A and E have degree 1. Second, the examples differ: Response~1 specifies \(P_4\) with formal vertex and edge sets, while Response~2 lists five lettered vertices. Third, Response~1 adds material absent from Response~2, including the general definition of \(P_n\), graph properties, applications, and related concepts.

\section{Human Annotation Study}
\label{app:human_annotation}

\subsection{Ethics and Annotator Information}
\label{app:human_ethics}

The study was administratively reviewed by the Institutional Review Board (IRB) and determined not to constitute Human Subjects Research under 45 CFR 46.102(e); therefore, IRB oversight was not required.

Eight fluent-English annotators were recruited via email from a university community and compensated with a \$25 gift card for completing both annotation rounds, which required approximately one hour in total. The annotation task involved judgments of AI-generated responses, not collecting information about the annotators themselves. No personally identifiable information about annotators is used in the analysis. 

\subsection{Response Pair Sampling and Construction}
\label{app:human_sampling}

We build the human-annotation set from a pool of 500 response pairs in \textsc{Infinity-Chat}. For each pair, we compute the cosine similarity between the two responses using \texttt{Qwen3-Embedding-8B} and stratify the pairs into quartiles of the resulting similarity distribution. To avoid extreme cases, we discard pairs below the 5th or above the 95th percentile, and then sample from all four quartiles so that the annotation set covers a broad range of embedding similarity.

The sampled pairs underrepresent several binary form variations. To improve their coverage, we edit a subset of the sampled responses, targeting headers, bullet points, opening templates, and closing templates, with edits designed to preserve the original semantic content. The resulting 90 pairs form the main set used throughout the human study.

\begin{table*}[t]
\centering
\small
\begin{tabularx}{\textwidth}{@{}l l >{\raggedright\arraybackslash}X >{\raggedright\arraybackslash}X@{}}
\toprule
Type & Per annotator & Source & Role \\
\midrule
Main        & 33 or 34       & 90 sampled pairs, each seen by 3 annotators                        & Primary judgments \\
Halo        & 3 (shared)     & Lowest-quartile similarity pairs outside the main set, with all form features closely matched & Testing whether surface-form similarity leads to inflated similarity ratings \\
Repeat      & 3              & Annotator's own main pairs, A/B reversed                           & Within-annotator consistency \\
Catch       & 2              & Randomly sampled responses, identical on both sides ($r_A = r_B$)  & Attention check \\
\midrule
Total       & 41 or 42       & & \\
\bottomrule
\end{tabularx}
\caption{Item types in the human annotation study.}
\label{tab:human_item_types}
\end{table*}

\subsection{Annotation Items and Assignment}
\label{app:human_distribution}
\label{app:human_assignment}

The human study comprises 90 \emph{main} response pairs and three types of diagnostic items: \emph{halo}, \emph{repeat}, and \emph{catch} pairs. Each annotator receives 33 or 34 main pairs, three shared halo pairs, three repeat pairs, and two catch pairs, for a total of 41 or 42 items. Table~\ref{tab:human_item_types} summarizes the item types.

\paragraph{Main Pairs.}
The 90 main pairs cover a broad range of semantic and form similarity. Each pair is assigned to exactly three annotators, yielding 270 pair assignments per round. Assignments are balanced across similarity ranges and form features.

\paragraph{Halo Pairs.}
All annotators evaluate the same three halo pairs, which are not included among the main pairs. These pairs are sampled from the lowest quartile of semantic similarity in \textsc{Infinity-Chat} (similarity in $[0.50, 0.67]$), a relatively low range for responses to the same prompt. We then manually edit both responses in each pair to include all five binary form features (headers, bullet points, opening templates, closing templates, and emphasis). Each halo pair thus combines low semantic similarity with identical binary form, allowing us to test whether shared surface form inflates perceived similarity despite semantic differences. 

\paragraph{Repeat Pairs.}
Each annotator sees three main pairs a second time, \emph{with the A/B order reversed} relative to their original presentation. Each repeat appears after its original, with at least 12 items in between. The three repeats are chosen to cover at least three similarity ranges. Repeat pairs are used to assess within-annotator consistency.

\paragraph{Catch Pairs.}
Each annotator receives two catch pairs in which the same response appears on both sides ($r_A = r_B$). These responses are drawn from main-set samples not otherwise assigned to that annotator, and the two catch items use distinct source prompts and responses.

\paragraph{Presentation Order.}
Items are ordered so that similarity ranges, repeated prompts and responses are not locally concentrated. Diagnostic items are distributed throughout the sequence, and the first and last items are always ordinary main pairs. 

\begin{figure*}[t]
    \centering
    \begin{subfigure}[b]{0.85\textwidth}
        \centering
        \includegraphics[width=\linewidth]{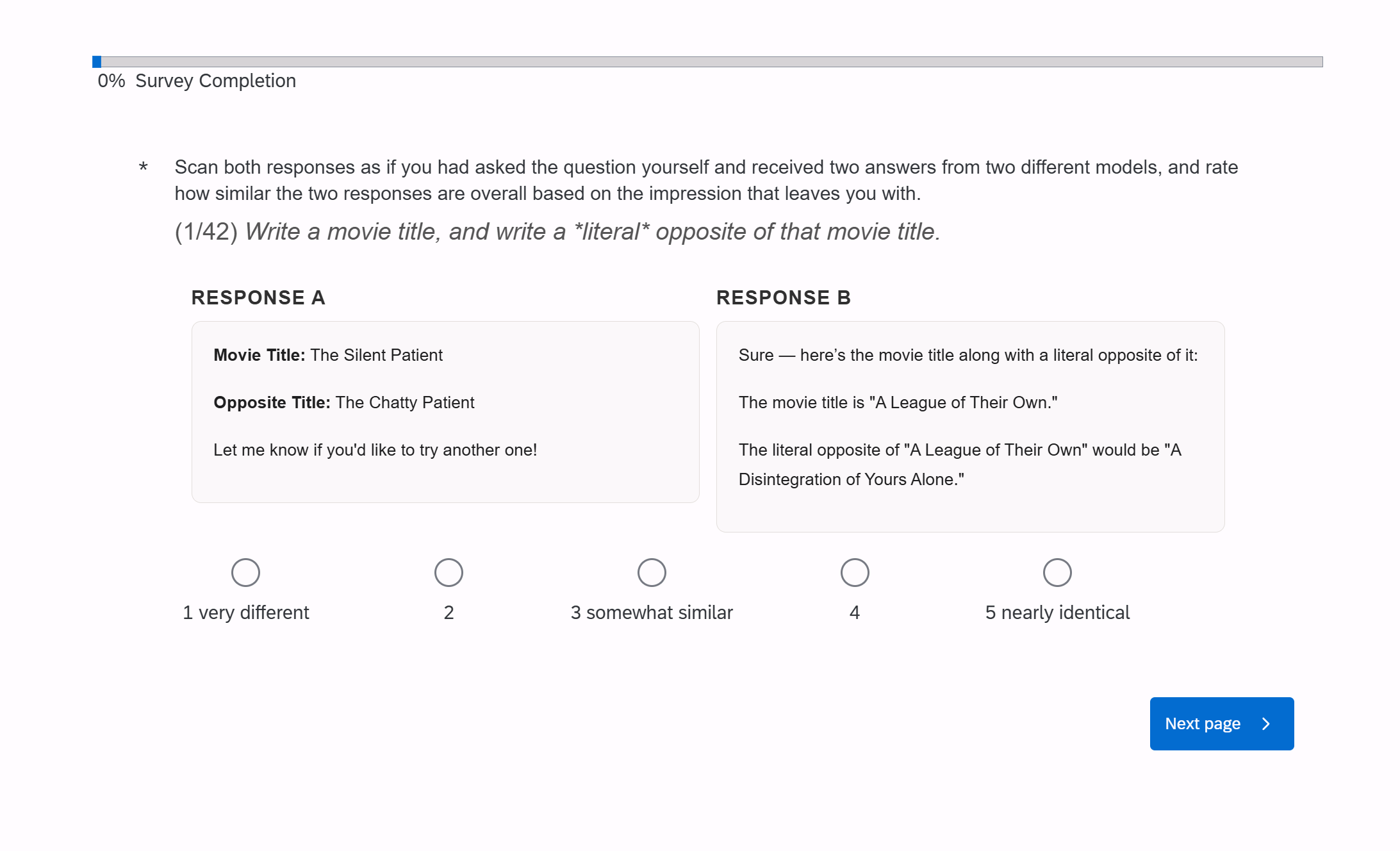}
        \caption{Round~1: overall similarity.}
        \label{fig:interface_r1}
    \end{subfigure}

    \vspace{1em}

    \begin{subfigure}[b]{0.85\textwidth}
        \centering
        \includegraphics[width=\linewidth]{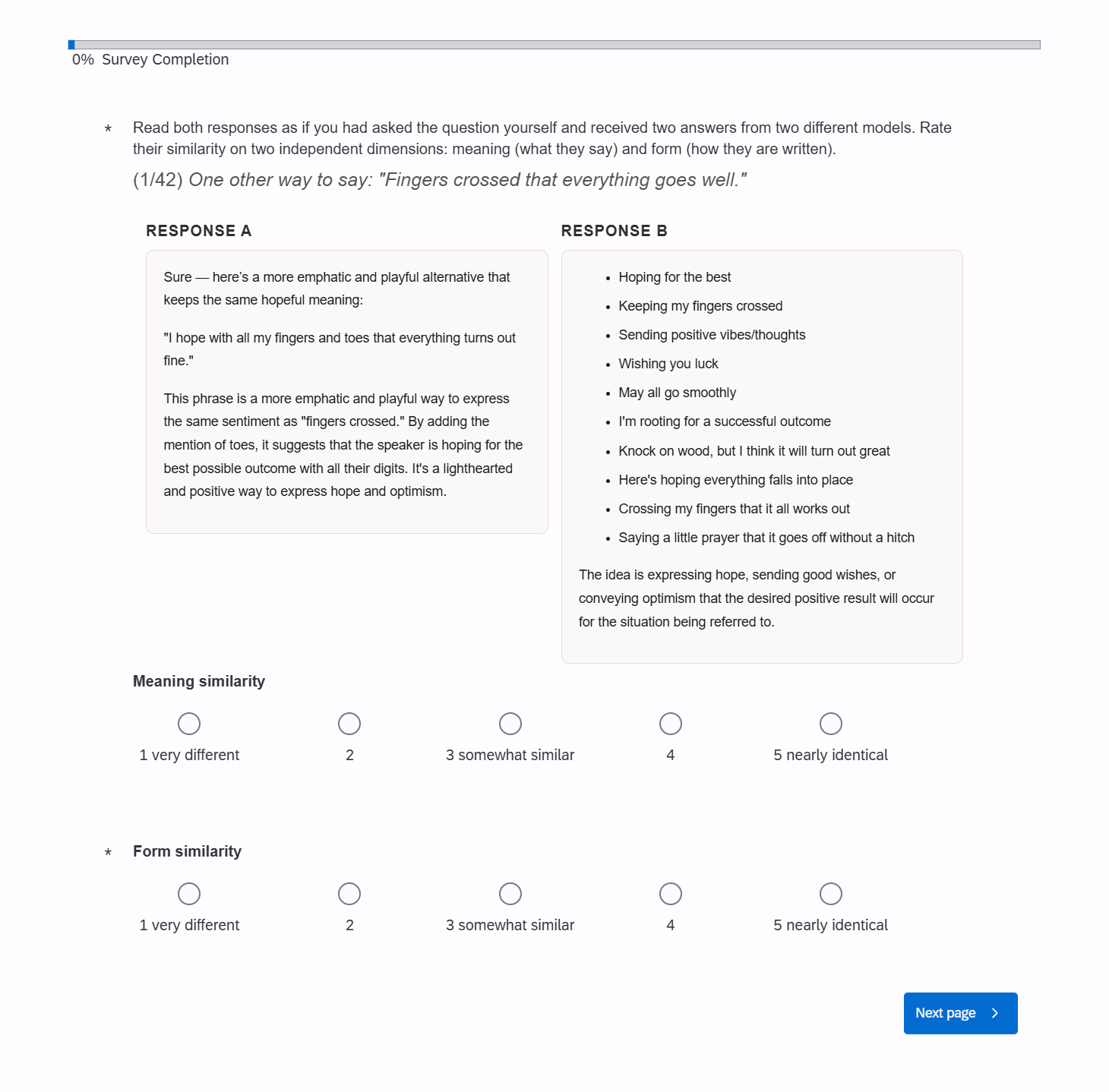}
        \caption{Round~2: meaning and form similarity.}
        \label{fig:interface_r2}
    \end{subfigure}

    \vspace{0.6em}
    \caption{Annotation interface on Qualtrics. Each item shows the question and two responses side by side, followed by a five-point scale with verbal labels at points 1, 3, and 5. In Round~2, annotators give separate meaning and form ratings on the same page.}
    \label{fig:annotation_interface}
\end{figure*}
\subsection{Annotation Protocol}
\label{app:human_protocol}

Annotation is conducted in two rounds on the same set of response pairs. In Round~1, annotators give a single overall similarity rating for each pair. In Round~2, they rate meaning similarity and form similarity separately. Instructions for each round are distributed only before that round, so Round~1 judgments are made without knowledge of the Round~2 criteria. Annotators are instructed to judge only how alike the two responses are, not which one is better or more correct, and to use the full range of the scale. Each set of instructions includes three worked examples, all answering the same question, that illustrate the rating criteria. These examples are not part of the annotation set. 

\paragraph{Round 1: Overall Similarity.}
Annotators give a single rating reflecting their overall impression of how similar the two responses are, reading them as if they had received two answers to their own question rather than comparing them point by point. The instructions mention content, visual presentation, wording, tone, and overall takeaway as aspects that may shape this impression, but ask annotators not to rate or average them separately. Ratings use a five-point scale. The written guidelines define all five points: 1 (\emph{very different}: as different as two answers to this question could be), 2 (\emph{clearly different}), 3 (\emph{somewhat similar}), 4 (\emph{clearly similar}), and 5 (\emph{nearly identical}). To keep the interface uncluttered, the rating scale itself displays verbal labels only at points 1, 3, and 5 (Figure~\ref{fig:annotation_interface}).

\paragraph{Round 2: Meaning and Form Similarity.}
Annotators rate the same pairs again on two independent dimensions, using the same scale. \emph{Meaning similarity} concerns what the responses say (main points, recommendations, and information conveyed), ignoring length, layout, and wording style. \emph{Form similarity} concerns how they are written (visual impression, structure and markup, sentence construction, and register), ignoring content. Annotators are told not to anchor on their Round~1 ratings.

Both rounds are administered through the Qualtrics survey platform, and Round~2 takes place one day after Round~1. Figure~\ref{fig:annotation_interface} shows the annotation interface used in each round.

\subsection{Annotator Agreement and Reliability}
\label{app:human_agreement}

We assess annotation reliability from three complementary perspectives: agreement across annotators on the main pairs, within-annotator consistency on repeated pairs, and performance on identical-response catch items.

\paragraph{Inter-Annotator Agreement.}
Because ratings are collected on an ordinal five-point scale and the assignment is incomplete across annotators, we report ordinal Krippendorff's $\alpha$ together with pairwise Spearman correlations between annotators who share main items. Agreement is highest for form similarity, with ordinal Krippendorff's $\alpha=0.606$, compared with $\alpha=0.373$ for overall similarity and $\alpha=0.329$ for meaning similarity. The corresponding pairwise Spearman correlations, weighted by the number of jointly annotated items, are $\rho=0.688$ for form, $\rho=0.467$ for overall, and $\rho=0.445$ for meaning. 

Given the five-point scale, we additionally examine near agreement. Across main pairs, two annotators' ratings differ by at most one point in 84.1\% of comparisons for form similarity, 78.1\% for overall similarity, and 81.5\% for meaning similarity. These results indicate that annotators are particularly consistent in distinguishing form, while overall and meaning judgments exhibit greater subjectivity.

\paragraph{Within-Annotator Consistency.}
Repeat pairs provide a direct test of whether annotators give stable judgments when seeing the same response pair again after a substantial delay and with the response order reversed. Across the 24 repeat judgments in each dimension, exact ratings are reproduced in 16/24 cases for overall similarity, 18/24 for form similarity, and 19/24 for meaning similarity. The mean absolute change is only 0.33, 0.25, and 0.21 points, respectively, on the five-point scale. Every repeat differs from its original rating by at most one point. This indicates high within-annotator stability despite the reversed presentation order.

\paragraph{Catch Items.}
Catch items test whether annotators recognize pairs in which the two responses are identical. In Round~1, all 16 catch judgments receive the maximum similarity score of 5. In Round~2, 15/16 catch judgments receive a 5 for both form and meaning similarity; the remaining judgment is a 4 in both dimensions. Thus, 46 of the 48 catch judgments across the two rounds receive the expected maximum score, with no catch judgment below 4.

\subsection{Illustrative Halo Pairs}
\label{app:human_halo}

Table~\ref{tab:halo_results} summarizes the ratings for the three halo pairs, together with ordinary lowest-quartile pairs as a reference.

\begin{table}[t]
\centering
\small
\begin{tabular}{lccc}
\toprule
Pair & Overall & Meaning & Form \\
\midrule
Lowest-quartile pairs (mean) & 2.13 & 2.62 & 2.72 \\
\midrule
H1 & 2.13 & 2.00 & 4.00 \\
H2 & 2.50 & 1.75 & 4.13 \\
H3 & 2.63 & 2.50 & 3.63 \\
\bottomrule
\end{tabular}
\caption{Mean human ratings for the three halo pairs H1--H3 ($n=8$ annotators per pair), with ordinary pairs from the lowest semantic-similarity quartile shown as a reference.}
\label{tab:halo_results}
\end{table}

These pairs provide directional evidence that form similarity influences overall similarity judgments: relative to ordinary lowest-quartile pairs, all three halo pairs receive higher form ratings and lower meaning ratings, yet their overall ratings are somewhat above the lowest-quartile mean.

\section{Models, Responses, Prompts, and Filtering}
\label{app:models_response}

\subsection{Model List and Computing Cost}

Table~\ref{tab:models} lists all evaluated models. Hybrid-reasoning models (Gemma~4, Qwen3-14B, Qwen3.5-9B) are evaluated in non-thinking mode. Our experiments used eight NVIDIA H100 GPUs, consuming about 2,000 GPU-hours in total. 

\begin{table*}[!t]
\centering
\small
\setlength{\tabcolsep}{6pt}
\renewcommand{\arraystretch}{1.05}
\begin{tabular}{@{}lllc@{}}
\toprule
\textbf{Model} & \textbf{Series} & \textbf{Stage} & \textbf{Ref.} \\
\midrule
\multicolumn{4}{@{}l}{\emph{API-served}} \\
\midrule
\texttt{deepseek-v4-flash-0423}         & deepseek-v4 & - & \cite{deepseekai2026deepseekv4} \\
\texttt{gemini-3.1-flash-lite}         & gemini-3.1  & - & \cite{googledeepmind2026gemini31flashlite} \\
\texttt{gpt-5.4-nano}                 & gpt-5.4     & - & \cite{openai2026gpt54mininano} \\
\midrule
\multicolumn{4}{@{}l}{\emph{Gemma}} \\
\midrule
\texttt{gemma-2-9b-it}                & gemma-2 & Instruct & \cite{gemma2} \\
\texttt{gemma-3-12b-it}               & gemma-3 & Instruct & \cite{gemma3} \\
\texttt{gemma-4-12B-it}               & gemma-4 & Instruct$^{\dagger}$ & \cite{gemma4} \\
\midrule
\multicolumn{4}{@{}l}{\emph{IBM Granite}} \\
\midrule
\texttt{granite-3.1-8b-instruct}       & granite-3.1 & Instruct & \cite{granite31} \\
\texttt{granite-3.3-8b-instruct}       & granite-3.3 & Instruct$^{\dagger}$ & \cite{granite33} \\
\texttt{granite-4.1-8b}               & granite-4.1 & Instruct & \cite{granite41} \\
\midrule
\multicolumn{4}{@{}l}{\emph{GLM}} \\
\midrule
\texttt{glm-4-9b-chat}                & glm-4   & Instruct & \cite{glm4} \\
\texttt{GLM-4-9B-0414}                & glm-4   & Instruct & \cite{glm40414} \\
\texttt{GLM-4.7-Flash}                & glm-4.7 & Instruct$^{\dagger}$ & \cite{glm47flash} \\
\midrule
\multicolumn{4}{@{}l}{\emph{OLMo 2 / OLMo 3 / OLMo Hybrid}} \\
\midrule
\texttt{OLMo-2-1124-7B-Instruct}       & olmo-2      & Instruct & \cite{olmo2} \\
\texttt{OLMo-3-1025-7B}               & olmo-3      & Base     & \cite{olmo3} \\
\texttt{OLMo-3-7B-Instruct-SFT}       & olmo-3      & SFT      & \\
\texttt{OLMo-3-7B-Instruct-DPO}       & olmo-3      & DPO      & \\
\texttt{OLMo-3-7B-Instruct}           & olmo-3      & RLVR     & \\
\texttt{OLMo-Hybrid-Instruct-DPO-7B}  & olmo-hybrid & DPO$^{\ddagger}$ & \cite{olmohybrid} \\
\midrule
\multicolumn{4}{@{}l}{\emph{Qwen 2.5 / Qwen 3 / Qwen 3.5}} \\
\midrule
\texttt{Qwen2.5-14B-Instruct}         & qwen2.5 & Instruct & \cite{qwen25} \\
\texttt{Qwen3-14B}                   & qwen3   & Instruct$^{\dagger}$ & \\
\texttt{Qwen3.5-9B}                  & qwen3.5 & Instruct$^{\dagger}$ & \cite{qwen35} \\
\midrule
\multicolumn{4}{@{}l}{\emph{T\"ulu 3 (Llama-3.1-8B base)}} \\
\midrule
\texttt{Llama-3.1-8B}                 & llama-3.1 & Base & \cite{llama3} \\
\texttt{Llama-3.1-Tulu-3-8B-SFT}      & tulu-3 & SFT  & \cite{tulu3} \\
\texttt{Llama-3.1-Tulu-3-8B-DPO}      & tulu-3 & DPO  & \\
\texttt{Llama-3.1-Tulu-3-8B-RLVR}     & tulu-3 & RLVR & \\
\bottomrule
\end{tabular}
\caption{Models evaluated in this work, grouped by series and training stage. \textbf{Base}: pretrained checkpoint; \textbf{SFT}: supervised fine-tuning; \textbf{DPO}: direct preference optimization; \textbf{RLVR}: reinforcement learning with verifiable rewards; \textbf{Instruct}: official instruction-tuned release; \textbf{Thinking}: release with explicit reasoning traces; \textbf{API-served}: models accessed through APIs. $^{\dagger}$Models supporting reasoning, evaluated in non-thinking mode. $^{\ddagger}$Only a DPO checkpoint has been released for this series, so we use it
for the temporal comparison.}
\label{tab:models}
\end{table*}

\begin{table*}[htbp]
\centering
\scriptsize 
\begin{tabular}{p{0.05\textwidth} p{0.90\textwidth}}
\hline
\textbf{ID} & \textbf{Question / Prompt} \\ \hline
\multicolumn{2}{l}{\textbf{Creative Content Generation}} \\ \hline
1 & Create a character named Bill Honda. \\
2 & Show me a joke? \\
3 & Write about puns with the word ``solana'' and animal names. \\
4 & Write a poem about a man. \\
5 & I once got fired from a keyboard factory for not putting in enough shifts. Make up a similar joke about ships. \\
6 & Generate a title about a penguin. \\
7 & Give me creative nicknames. \\
8 & Generate an image prompt for `inside a Lamborghini full of puppies and kittens.' \\
9 & Write about a pink fluffy alien. \\
10 & Generate prompts to create space exploration coloring book images for toddlers. \\ \hline
\multicolumn{2}{l}{\textbf{Brainstorm \& Ideation}} \\ \hline
11 & Suggest a theme for shoe care. \\
12 & Can you imagine fictional cable TV networks inspired by Bravo in 1982, with details including viewership? \\
13 & Come up with a simple idea for a pixel game in the arcade genre. \\
14 & Please give some ideas for unconventional `railways' to model. They can include monorails and cableways. \\
15 & Give me ideas for academic research on the topic of SEO. \\
16 & Give me content for a LinkedIn post about the top reasons why prototyping first could save your digital product. \\
17 & Suggest ideas for visualization with four different visual channels (also known as visual variables). \\
18 & Give me Teacher's Day gift ideas. \\
19 & How to create a roadmap for an accessibility center of excellence? \\
20 & Write captions related to butterflies without using the word `butterfly'. \\ \hline
\multicolumn{2}{l}{\textbf{Open-Endedness}} \\ \hline
21 & What are some things that the 1980s had in common with the 1960s, but the 1990s did not have in common? \\
22 & What specs would a premium laptop from 1997, 1999, and 2002 have in detail (plus the price)? \\
23 & If one day suddenly became 48 hours, would humans be able to adapt to it? \\
24 & How would a computer from 2010 perform in comparison to computers from 1960 to 2060? \\
25 & What changes will happen if a TV that delivers smells is developed? \\
26 & If you could see five minutes into the future, how would certain sports play out? \\
27 & Evaluate the statement: `Sentience is the ability to feel pain.' \\
28 & What will happen if we have unlimited electricity for free? \\
29 & Whether artificial intelligence can actually create something truly new. \\
30 & Can you try to objectively rank which pizzas are the best ones? \\ \hline
\multicolumn{2}{l}{\textbf{Alternative Perspectives}} \\ \hline
31 & Is being a game QA tester a good job? \\
32 & Do you agree that athletes' salaries should be proportional to the sport's popularity? \\
33 & What is the best feeling in the world? \\
34 & I think, if a software engineer says something like ``it's not a bug, it's a feature,'' they're a bad software engineer. \\
35 & Is it true that it's easier to read physical books and write physical notes compared to their digital counterparts? \\
36 & How do you feel about most of your college classes happening virtually? Discuss what you like and dislike about online classes. \\
37 & Discuss whether antivirus software should be paid for, considering that people often encounter viruses and not everyone has that kind of money. \\
38 & Are static types better than dynamic types? \\
39 & What are some best ways to make a lot of money while also making a positive impact on the world? \\
40 & Linux vs. Windows. Why Linux? What is better or special about Linux that differentiates it from Windows? \\ \hline
\multicolumn{2}{l}{\textbf{Alternative Styles}} \\ \hline
41 & What are good informal and formal responses to `see you when I see you'? \\
42 & How to say thank you for your time in the subject line of an email? \\
43 & Write a professional greeting message for my friend, Mr. Qubow. \\
44 & Reply to an email that once I receive feedback from the store, I will reach back to you. \\
45 & Draft a follow-up email to an interviewer about setting up a time to speak. \\
46 & I want to ask my sister if I can borrow her slime. Can you help me write a note? Her name is Lily. \\
47 & Make a story about adorable Kirby going about his day. \\
48 & Write a story about Belle, a pink, antennaed alien girl. She has the power to melt into sludge. \\
49 & Kindly inform the client that the standard purchase has been initiated, but according to the person in charge of the purchase, this may take some time and we are awaiting their response. \\
50 & Write tweets about another video from today's sunrise on the beach. \\ \hline
\multicolumn{2}{l}{\textbf{Information-Seeking}} \\ \hline
51 & How to manage a classroom effectively? \\
52 & How do I perform a literature review for my research paper? \\
53 & Tell me about the assembly language of a computer. \\
54 & What are the different types of shoes? \\
55 & How do I set up a website, from domain purchase to uploading my site? \\
56 & Can you give me recipes for 4 people? My budget is 100 euros. \\
57 & Which easy dinner meals would you recommend to a person who doesn't like to cook? \\
58 & Give me a quick way to remember how to prove a mathematical proposition. \\
59 & Show an example of how to use threading in Python. \\
60 & Explain everything about using garlic in Indian cooking from first principles. \\ \hline
\end{tabular}
\caption{Example questions by category. Questions are sampled from \textsc{Infinity-Chat} \citep{jiang2025artificial}; some are edited to remove stylistic constraints and format requirements.}
\label{tab:questions}
\end{table*}

\subsection{Prompts and Question Types}
We sample open-ended questions from {\textsc{Infinity-Chat}} across six top-level categories: Creative Content Generation, Brainstorming and Ideation, Open-Endedness, Alternative Perspectives, Alternative Styles, and Information-Seeking. Questions with multiple top-level categories are assigned to a single representative one. We manually review the sampled questions for suitability and use GPT-5.4-Mini to remove stylistic constraints (e.g., ``in Hemingway's style'') and format requirements (e.g., ``write in bullet points'') that could bias evaluation. The final set contains 300 questions, 50 per category (examples in Table~\ref{tab:questions}).

\subsection{Experimental Settings for Stage-wise Analysis}
\label{app:settings_for_stage}
\paragraph{Models and stages.}
We study OLMo~3 (7B)~\citep{olmo3} and T\"ulu~3 (8B)~\citep{tulu3}, selected for their comparable model sizes and publicly available checkpoints at each post-training stage (Base, SFT, DPO, and RLVR). We use the same prompt pool and decoding configuration as in \S\ref{sec:temporal}. After filtering, 263 prompts remain for analysis.

\paragraph{Prompting base models.}
Because base models are not instruction-tuned and tend to be unstable when answering questions directly, we elicit their responses with three-shot prompting. Since demonstrations also carry formatting conventions, we consider two conditions: \textit{form}, which retains the original formatting markers (e.g., lists and Markdown emphasis), and \textit{pure}, which strips these markers while preserving the content. This design allows us to examine the effects of formatting cues while keeping the demonstration content approximately fixed. Note that \textit{pure} and \textit{form} are alternative prompting conditions for the same base checkpoint, not successive training stages. A short example is shown below.

\newtcolorbox{fewshotbox}[1]{%
  enhanced, breakable,
  colback=gray!3, colframe=gray!55,
  title={#1},
  boxrule=0.4pt,
  left=5pt, right=5pt, top=4pt, bottom=4pt,
  before skip=4pt, after skip=4pt
}

\begin{fewshotbox}{Example (\textit{form} variant)}
\small
\textbf{Input:} Write a movie title, and write a *literal* opposite 
of that movie title.

\vspace{2pt}
\textbf{Response:} **Movie Title:** "Fast and Furious" \quad 
**Literal Opposite:** "Slow and Calm".
\end{fewshotbox}

\begin{fewshotbox}{Example (\textit{pure} variant)}
\small
\textbf{Input:} Write a movie title, and write a literal opposite of 
that movie title.

\vspace{2pt}
\textbf{Response:} Movie Title: "Fast and Furious". Literal 
Opposite: "Slow and Calm".
\end{fewshotbox}

\paragraph{Distance computation.}
At each stage, we compute the inter-model distance between corresponding OLMo~3 and T\"ulu~3 checkpoints (e.g., SFT vs.\ SFT), alongside the intra-model distance within each checkpoint.

\subsection{Filtering Responses}
\label{app:filtering_prompts}

\citet{shypula2025evaluating} argue that small models often generate incoherent outputs, thereby inflating apparent diversity in evaluation, since random tokens are meaningless and would not be produced by more sophisticated models. While they use a code generation task, in which the validity of outputs can be checked directly, we use a text generation task, which is easier for smaller models. The notion of quality we refer to here is that the model does not generate meaningless content, which we further expand into three criteria and evaluate using Qwen-3.8-27b \citep{qwen38}: \textbf{relevance} (directly addresses the question; refusals fail), \textbf{completeness} (coherent and finished, with no truncation or repeated junk tokens), and \textbf{focus} (contains only the answer, with no metadata, debug text, or reasoning traces). A response is discarded if it fails any one of these. The exact prompt is reproduced below. 

\begin{tcolorbox}[
    enhanced, breakable,
    colback=gray!3, colframe=gray!55,
    title={Filtering prompt},
    left=5pt, right=5pt, top=4pt, bottom=4pt,
    before upper={\setlength{\parskip}{4pt}},
]
\small
\#\# Task

Given a QUESTION and a RESPONSE, check the RESPONSE against three checks. Do NOT judge quality, accuracy, length, detail, or whether every requirement or constraint in the QUESTION is met. Default to PASS; fail a check only when a listed condition clearly applies. Formatting (markdown, lists, labels, quotes, escaped characters like \texttt{\textbackslash"} or \texttt{\textbackslash n}) never causes a failure.

\#\# Checks

\begin{itemize}[label={-}, leftmargin=1em, itemsep=3pt, topsep=0pt]
    \item \textbf{Relevance}: fails only if the RESPONSE refuses the task, is about a completely different topic, or argues the opposite of what the QUESTION explicitly asks (watch for ``not'', ``against'', ``without'').
    \item \textbf{Completeness}: fails only if the RESPONSE is cut off mid-sentence or mid-list, is only a stub (e.g., just ``Dear recipient,''), or contains junk (e.g., \texttt{<br><br><br>}, 5+ blank lines, the same sentence repeated).
    \item \textbf{Focus}: fails only if the RESPONSE contains leaked text not meant for the user: model/system metadata, logs, token usage, tool-call traces, or reasoning markers (e.g., \texttt{<think>}, ``Thought:''). Intros, closings, tips, disclaimers, and extra detail are fine.
\end{itemize}
\end{tcolorbox}

Table~\ref{tab:filter_rate} reports the fraction of responses removed for each model. Filter rates are low for post-trained models (mostly below 4\%) but considerably higher for base models, especially when the few-shot examples contain no formatting markers (up to 24\%). Because filtering removes responses unevenly across models and prompts, every analysis is restricted to prompts for which \emph{each} model in the comparison retains at least 30 valid responses; the resulting number of prompts is given in the header of each group in Table~\ref{tab:filter_rate}.

\begin{table}[!t]
\centering
\scriptsize
\setlength{\tabcolsep}{6pt}
\renewcommand{\arraystretch}{1.05}
\begin{tabular}{@{}lr@{}}
\toprule
\textbf{Model} & \textbf{Filter rate (\%)} \\
\midrule
\multicolumn{2}{@{}l}{\emph{Cross-model \& temporal} \quad (18 models, 279 prompts)} \\
\midrule
\texttt{gemma-2-9b-it}                       &  3.46 \\
\texttt{gemma-3-12b-it}                      &  0.93 \\
\texttt{gemma-4-12B-it}$^{\dagger}$           &  0.91 \\
\addlinespace
\texttt{OLMo-2-1124-7B-Instruct}             &  2.71 \\
\texttt{Olmo-3-7B-Instruct}                  &  1.53 \\
\texttt{Olmo-Hybrid-Instruct-DPO-7B}         &  1.76 \\
\addlinespace
\texttt{Qwen2.5-14B-Instruct}                &  1.46 \\
\texttt{Qwen3-14B}$^{\dagger}$               &  0.86 \\
\texttt{Qwen3.5-9B}$^{\dagger}$              &  2.16 \\
\addlinespace
\texttt{glm-4-9b-chat}                       &  2.39 \\
\texttt{GLM-4-9B-0414}                       &  2.07 \\
\texttt{GLM-4.7-Flash}                       &  1.76 \\
\addlinespace
\texttt{granite-3.1-8b-instruct}             &  2.49 \\
\texttt{granite-3.3-8b-instruct}             &  2.21 \\
\texttt{granite-4.1-8b}                      &  1.41 \\
\addlinespace
\texttt{deepseek-v4-flash-0423}                   &  1.36 \\
\texttt{gemini-3.1-flash-lite}               &  0.66 \\
\texttt{gpt-5.4-nano}                        &  1.83 \\
\midrule
\multicolumn{2}{@{}l}{\emph{Training stage} \quad (10 models, 263 prompts)} \\
\midrule
\texttt{Olmo-3-1025-7B} (pure)               & 23.89 \\
\texttt{Olmo-3-1025-7B} (form)               &  8.99 \\
\texttt{Olmo-3-7B-Instruct-SFT}              &  2.86 \\
\texttt{Olmo-3-7B-Instruct-DPO}              &  2.41 \\
\texttt{Olmo-3-7B-Instruct}                  &  1.53 \\
\addlinespace
\texttt{Llama-3.1-8B} (pure)                 & 17.71 \\
\texttt{Llama-3.1-8B} (form)                 &  8.96 \\
\texttt{Llama-3.1-Tulu-3-8B-SFT}             &  5.67 \\
\texttt{Llama-3.1-Tulu-3-8B-DPO}             &  1.87 \\
\texttt{Llama-3.1-Tulu-3-8B-RLVR}            &  1.73 \\
\bottomrule
\end{tabular}
\caption{Percentage of responses discarded by the filter, grouped by analysis. All responses are sampled with top-$p$ sampling. For base models, \emph{pure} denotes few-shot prompting without formatting markers and \emph{form} denotes few-shot prompting with formatting markers. Group headers list the model count and the number of prompts (of 300) with $\geq$30 valid post-filter responses for all models. $^{\dagger}$Hybrid-reasoning model evaluated in non-thinking mode.}
\label{tab:filter_rate}
\end{table}

\begin{figure*}[!t]
    \centering
    \includegraphics[width=\textwidth]{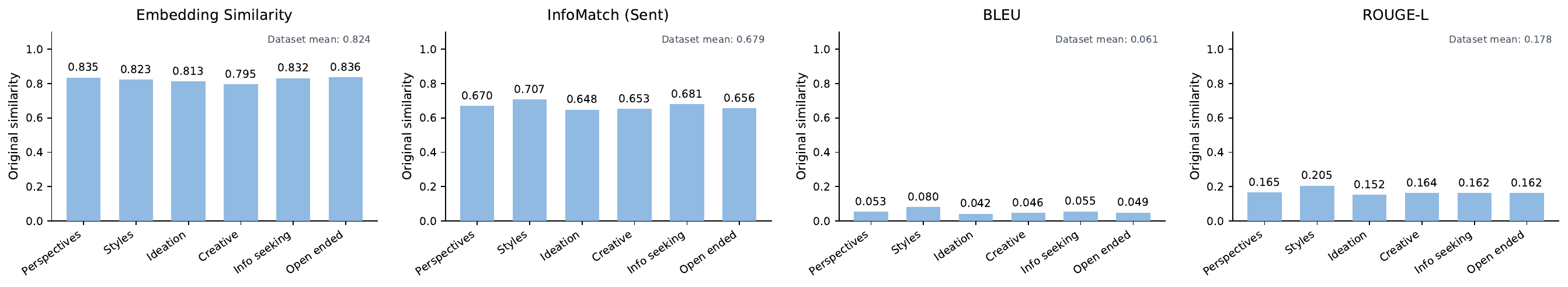}
\caption{Shared semantics with divergent wording holds across question types. Inter-model similarity by question type for 18 models; bars show the mean over all 153 model pairs. Questions with multiple labels count toward each assigned category. Upper-right values give each metric's mean over all 279 questions. Abbreviations: Perspectives/Styles = Alternative Perspectives/Styles; Ideation = Brainstorming and Ideation; Creative = Creative Content Generation; Info seeking = Information-Seeking; Open ended = Open-Endedness.}
\label{fig:question_type}
\end{figure*}

\definecolor{meaningbg}{HTML}{DCECF8}
\definecolor{formbg}{HTML}{F8E9D7}
\begin{table*}[t]
\centering
\small
\setlength{\tabcolsep}{4.5pt}
\renewcommand{\arraystretch}{1.13}
\ifdefined\numw\else\newlength{\numw}\fi
\ifdefined\starw\else\newlength{\starw}\fi
\ifdefined\ciw\else\newlength{\ciw}\fi
\settowidth{\numw}{$\mathbf{-0.000}$}
\settowidth{\starw}{$^{***}$}
\settowidth{\ciw}{\fontsize{6.8}{7.2}\selectfont$[-0.000,-0.000]$}
\newcommand{\rhoci}[4][]{%
  \makebox[\numw][r]{$#2$}%
  \makebox[\starw][l]{$^{#1}$}%
  \hspace{0.35em}%
  \raisebox{-0.45ex}{\makebox[\ciw][l]{\textcolor{gray}{\fontsize{6.8}{7.2}\selectfont$[#3,#4]$}}}}
\newcommand{\bestci}[4][]{\rhoci[#1]{\mathbf{#2}}{#3}{#4}}
\begin{tabular}{@{}lccc@{}}
\toprule
Metric & Overall & Meaning & Form \\
\midrule
\multicolumn{4}{l}{\textit{Semantic}} \\
\quad Embedding Similarity & \rhoci[***]{0.400}{0.209}{0.561} & \cellcolor{meaningbg}\rhoci[***]{0.417}{0.220}{0.586} & \rhoci{0.140}{-0.073}{0.340} \\
\quad InfoMatch (Sent) & \rhoci[***]{0.593}{0.450}{0.705} & \cellcolor{meaningbg}\bestci[***]{0.606}{0.462}{0.722} & \rhoci[*]{0.263}{0.059}{0.448} \\
\quad BLEU & \bestci[***]{0.608}{0.461}{0.722} & \cellcolor{meaningbg}\rhoci[***]{0.538}{0.358}{0.688} & \rhoci[**]{0.325}{0.104}{0.514} \\
\quad ROUGE-L & \rhoci[***]{0.581}{0.422}{0.707} & \cellcolor{meaningbg}\rhoci[***]{0.525}{0.345}{0.670} & \rhoci[*]{0.260}{0.035}{0.460} \\
\midrule
\multicolumn{4}{l}{\textit{Lexical}} \\
\quad Lexical Diversity (MTLD)$^\ddagger$ & \rhoci{0.126}{-0.121}{0.356} & \rhoci{0.111}{-0.120}{0.330} & \rhoci{0.064}{-0.159}{0.279} \\
\quad Word Frequency (Zipf) & \rhoci{0.023}{-0.195}{0.240} & \rhoci{-0.026}{-0.246}{0.186} & \rhoci{0.166}{-0.054}{0.369} \\
\midrule
\multicolumn{4}{l}{\textit{Syntactic}} \\
\quad POS Distribution & \rhoci[***]{0.349}{0.158}{0.518} & \rhoci[**]{0.291}{0.092}{0.476} & \rhoci[**]{0.274}{0.070}{0.461} \\
\quad Dependency Distribution & \rhoci[***]{0.466}{0.285}{0.615} & \rhoci[***]{0.377}{0.168}{0.560} & \rhoci{0.175}{-0.042}{0.373} \\
\quad Function Word Profile & \rhoci[*]{0.244}{0.027}{0.438} & \rhoci{0.133}{-0.093}{0.351} & \rhoci[*]{0.246}{0.039}{0.436} \\
\quad Function Word Density & \rhoci{0.100}{-0.114}{0.304} & \rhoci{0.043}{-0.175}{0.254} & \rhoci{0.206}{0.007}{0.389} \\
\midrule
\multicolumn{4}{l}{\textit{Structural}} \\
\quad Response Length (chars) & \rhoci{0.196}{-0.012}{0.386} & \rhoci[*]{0.217}{0.007}{0.411} & \cellcolor{formbg}\rhoci[***]{0.412}{0.214}{0.581} \\
\quad Paragraph Density & \rhoci{0.194}{-0.008}{0.379} & \rhoci[*]{0.227}{0.013}{0.426} & \cellcolor{formbg}\rhoci[***]{0.388}{0.203}{0.548} \\
\quad Header Use & \rhoci{0.085}{-0.111}{0.280} & \rhoci{0.108}{-0.086}{0.296} & \cellcolor{formbg}\rhoci[***]{0.389}{0.221}{0.547} \\
\quad List Use & \rhoci{0.205}{0.006}{0.391} & \rhoci{0.161}{-0.053}{0.365} & \cellcolor{formbg}\bestci[***]{0.590}{0.458}{0.708} \\
\quad List Density & \rhoci{0.199}{0.003}{0.382} & \rhoci{0.154}{-0.061}{0.358} & \cellcolor{formbg}\rhoci[***]{0.548}{0.393}{0.677} \\
\midrule
\multicolumn{4}{l}{\textit{Stylistic \& Presentational}} \\
\quad Readability & \rhoci{0.076}{-0.147}{0.297} & \rhoci{0.180}{-0.036}{0.389} & \rhoci{0.047}{-0.166}{0.259} \\
\quad Concreteness & \rhoci[***]{0.349}{0.146}{0.534} & \rhoci{0.157}{-0.050}{0.355} & \rhoci{0.098}{-0.100}{0.289} \\
\quad Emotion Profile & \rhoci[***]{0.376}{0.185}{0.545} & \rhoci[***]{0.363}{0.166}{0.528} & \rhoci{0.095}{-0.124}{0.305} \\
\quad Emotion Intensity & \rhoci{0.126}{-0.077}{0.320} & \rhoci{0.165}{-0.041}{0.361} & \rhoci{0.137}{-0.082}{0.342} \\
\quad Punctuation Profile & \rhoci[***]{0.547}{0.392}{0.672} & \cellcolor{meaningbg}\rhoci[***]{0.485}{0.314}{0.624} & \rhoci[**]{0.309}{0.107}{0.487} \\
\quad Punctuation Density & \rhoci[*]{0.254}{0.057}{0.429} & \rhoci[*]{0.224}{0.024}{0.412} & \rhoci[*]{0.225}{0.011}{0.424} \\
\quad Opening Template & \rhoci{0.069}{-0.129}{0.269} & \rhoci{0.031}{-0.171}{0.231} & \rhoci{-0.007}{-0.196}{0.188} \\
\quad Closing Template & \rhoci{0.150}{-0.060}{0.354} & \rhoci{0.100}{-0.112}{0.307} & \rhoci{0.028}{-0.178}{0.234} \\
\quad Markdown Emphasis Use & \rhoci{0.157}{-0.043}{0.360} & \rhoci{0.197}{0.000}{0.383} & \rhoci[*]{0.270}{0.068}{0.454} \\
\quad Emphasis Density & \rhoci[*]{0.262}{0.064}{0.446} & \rhoci[**]{0.321}{0.125}{0.501} & \rhoci[**]{0.277}{0.069}{0.459} \\
\midrule
\multicolumn{4}{l}{\textit{Style embedding (reference)}} \\
\quad StyleDistance & \rhoci[***]{0.554}{0.398}{0.675} & \rhoci[***]{0.415}{0.238}{0.569} & \rhoci[***]{0.509}{0.329}{0.657} \\
\bottomrule
\end{tabular}
\caption{Spearman correlations between each metric and human ratings ($n=90$ response pairs). Smaller gray values show 95\% percentile bootstrap CIs (10,000 resamples of 90 pairs with replacement). Blue and orange cells mark the five metrics with the highest observed correlations with human meaning and form judgments, respectively. These correspond to the highlighted metrics in Table~\ref{tab:current_models} and the groups in Table~\ref{tab:temporal_grouped}. Bold marks each column's highest correlation. All metrics are sign-aligned. Stars denote unadjusted Spearman $p$-values: $^{*}p<0.05$, $^{**}p<0.01$, $^{***}p<0.001$. StyleDistance is a reference metric outside our inventory. $^\ddagger$MTLD uses $n=75$ valid pairs, with undefined values excluded in each bootstrap resample.}
\label{tab:human_metric_corr}
\end{table*}

\section{Correlation}
\subsection{Robustness to Embedding Models}
\label{app:embedding_robustness}

To verify that the patterns reported in Section~\ref{sec:compared_with_embedding} are not specific to Qwen3-Embedding-8B~\citep{qwen3embedding}, we repeat the analysis on the 500 response pairs of \textsc{Infinity-Chat} with two alternative embedding models, jina-embeddings-v5-text~\citep{akram2026jina} and QZhou-Embedding~\citep{yu2025qzhouembeddingtechnicalreport}. For each model, we compute the Spearman correlation between its whole-response embedding similarity and every metric in our framework. Because InfoMatch relies on sentence representations from the same encoder, it is recomputed with the corresponding embedding model. As Table~\ref{tab:embedding_robustness} shows, the correlations are largely consistent in direction and magnitude across the three embedding models.

\begin{table*}[t]
\centering
\small
\resizebox{\textwidth}{!}{%
\renewcommand{\arraystretch}{1.2}
\begin{tabular}{@{}llp{5.7cm}llc@{}}
\toprule
\textbf{Dimension} & \textbf{Metric} & \textbf{Description} & \textbf{Output} & \textbf{Similarity} & \textbf{Ref.} \\
\midrule

\multirow{4}{*}{Semantic}
& Embedding Similarity
& Cosine similarity between Qwen3-Embedding-8B representations of the complete responses; used as the primary semantic baseline
& pairwise & native & \cite{qwen3embedding} \\

& InfoMatch (Sent)
& Sentence-level semantic correspondence computed by bidirectional best-match cosine matching between Qwen3-Embedding-8B sentence representations, with the two directional scores combined using their harmonic mean; code blocks and tables are excluded
& pairwise & native & \cite{zhang2020bertscore,qwen3embedding} \\

& BLEU
& Symmetric response-level BLEU using SacreBLEU sentence scoring with lowercased up-to-4-gram overlap, effective order, exponential smoothing, and brevity penalty; code blocks are excluded
& pairwise & native & \cite{papineni2002bleu} \\

& ROUGE-L
& F1 score based on the longest common subsequence of the normalized textual content; code blocks are excluded
& pairwise & native & \cite{lin2004rouge} \\

\midrule

\multirow{2}{*}{Lexical}
& Lexical Diversity (MTLD)
& Length-robust measure of vocabulary diversity within a response, computed for responses containing at least 50 lexical tokens
& scalar & $-|\Delta|$ & \cite{mccarthy2010mtld} \\

& Word Frequency (Zipf)
& Mean Zipf frequency of words in a response according to \texttt{wordfreq}; out-of-vocabulary words receive frequency 0 and remain in the average
& scalar & $-|\Delta|$ & \cite{robyn_speer_2022_7199437,vanheuven2014subtlex} \\

\midrule

\multirow{4}{*}{Syntactic}
& POS Distribution
& Distribution over spaCy coarse part-of-speech categories among non-punctuation tokens, characterizing grammatical composition
& distribution & $-\mathrm{JSD}$ & \cite{honnibal2020spacy} \\

& Dependency Distribution
& Distribution over spaCy dependency relations among non-punctuation tokens, characterizing recurring syntactic relations
& distribution & $-\mathrm{JSD}$ & \cite{honnibal2020spacy} \\

& Function Word Profile
& Distribution across seven LIWC-style function-word classes (article / preposition / cconj / sconj / pronoun / auxiliary / particle) identified using POS-based rules with lexical matching for articles
& distribution & $-\mathrm{JSD}$ & \cite{pennebaker2015liwc,honnibal2020spacy} \\

& Function Word Density
& Number of detected function words per 100 lexical words
& scalar & $-|\Delta|$ & \cite{pennebaker2015liwc,honnibal2020spacy} \\

\midrule

\multirow{5}{*}{Structural}
& Response Length (chars)
& Total length of the original response in characters, capturing overall response extent
& scalar & $-|\Delta|$ & \\

& Paragraph Density
& Number of structural paragraphs per 100 prose lexical words; fenced code blocks are collapsed to single structural blocks before paragraph counting
& scalar & $-|\Delta|$ & \\

& Header Use
& Whether the response contains at least one explicit Markdown heading, detected from the parsed Markdown structure
& binary & $\mathbb{I}[=]$ & \\

& List Use
& Whether the response contains at least one Markdown list item or supported nonstandard bullet-marked line, detected from the parsed response structure
& binary & $\mathbb{I}[=]$ & \\

& List Density
& Number of list items per 100 lexical words
& scalar & $-|\Delta|$ & \\

\midrule

\multirow{10}{*}{\shortstack[l]{Stylistic \&\\Presentational}}
& Readability
& Flesch--Kincaid grade level computed from prose content, with sentence boundaries obtained by parsing each prose block independently
& scalar & $-|\Delta|$ & \cite{kincaid1975flesch} \\

& Concreteness
& Mean predicted word concreteness score (1--5; FastText regression), averaged over words with available FastText representations
& scalar & $-|\Delta|$ & \cite{brysbaert2014concreteness,bojanowski2017fasttext} \\

& Emotion Profile
& Normalized distribution over 28 GoEmotions category scores after subword-weighted aggregation across response chunks, characterizing the relative composition of affective expression
& distribution & $-\mathrm{JSD}$ & \cite{demszky2020goemotions} \\

& Emotion Intensity
& Maximum non-neutral GoEmotions probability after subword-weighted aggregation across response chunks, characterizing the strength of affective expression
& scalar & $-|\Delta|$ & \cite{demszky2020goemotions} \\

& Punctuation Profile
& Distribution over 11 punctuation classes: comma, period, dash (em dash, non-numeric-range en dash, \texttt{--}, or spaced hyphen), hyphen, semicolon, colon, exclamation mark, question mark, ellipsis, parentheses, and quotation marks
& distribution & $-\mathrm{JSD}$ & \\

& Punctuation Density
& Number of punctuation marks per 100 lexical words
& scalar & $-|\Delta|$ & \\

& Opening Template
& Whether the response begins with an interactional opener acknowledging or affirming the query, detected by an LLM-based classifier
& binary & $\mathbb{I}[=]$ & \\

& Closing Template
& Whether the response ends with an interactional closing or invitation for further engagement, detected by an LLM-based classifier
& binary & $\mathbb{I}[=]$ & \\

& Markdown Emphasis Use
& Whether the response contains at least one bold or italic span, detected from the parsed Markdown structure
& binary & $\mathbb{I}[=]$ & \\

& Emphasis Density
& Number of Markdown bold or italic spans per 100 lexical words
& scalar & $-|\Delta|$ & \\

\bottomrule
\end{tabular}%
}
\caption{Metric inventory for the five dimensions of LLM response similarity, details can be found in Appendix~\ref{app:metric_implementation}.}
\label{tab:framework_metrics}
\end{table*}

\subsection{Correlation with Human Similarity}
\label{app:human_check}
Section~\ref{sec:human_similarity} summarizes the overall tendencies of how response-similarity metrics relate to human judgments. Table~\ref{tab:human_metric_corr} reports the corresponding values: for each metric, the Spearman correlation between its pairwise similarity score and the human ratings of overall, meaning, and form similarity on the 90 main response pairs. As a reference, we additionally include StyleDistance~\citep{patel2025styledistance}, an embedding model trained specifically to capture writing style. Like Embedding Similarity, however, it summarizes a response pair with a single score and therefore offers no indication of which textual properties make two responses alike or different.

\section{Supplementary Results}
\subsection{Full Distance Trajectories}
\label{app:full_trajectories}
\label{app:stage_full}

\begin{figure*}[!t]
    \centering
    \includegraphics[width=\textwidth]{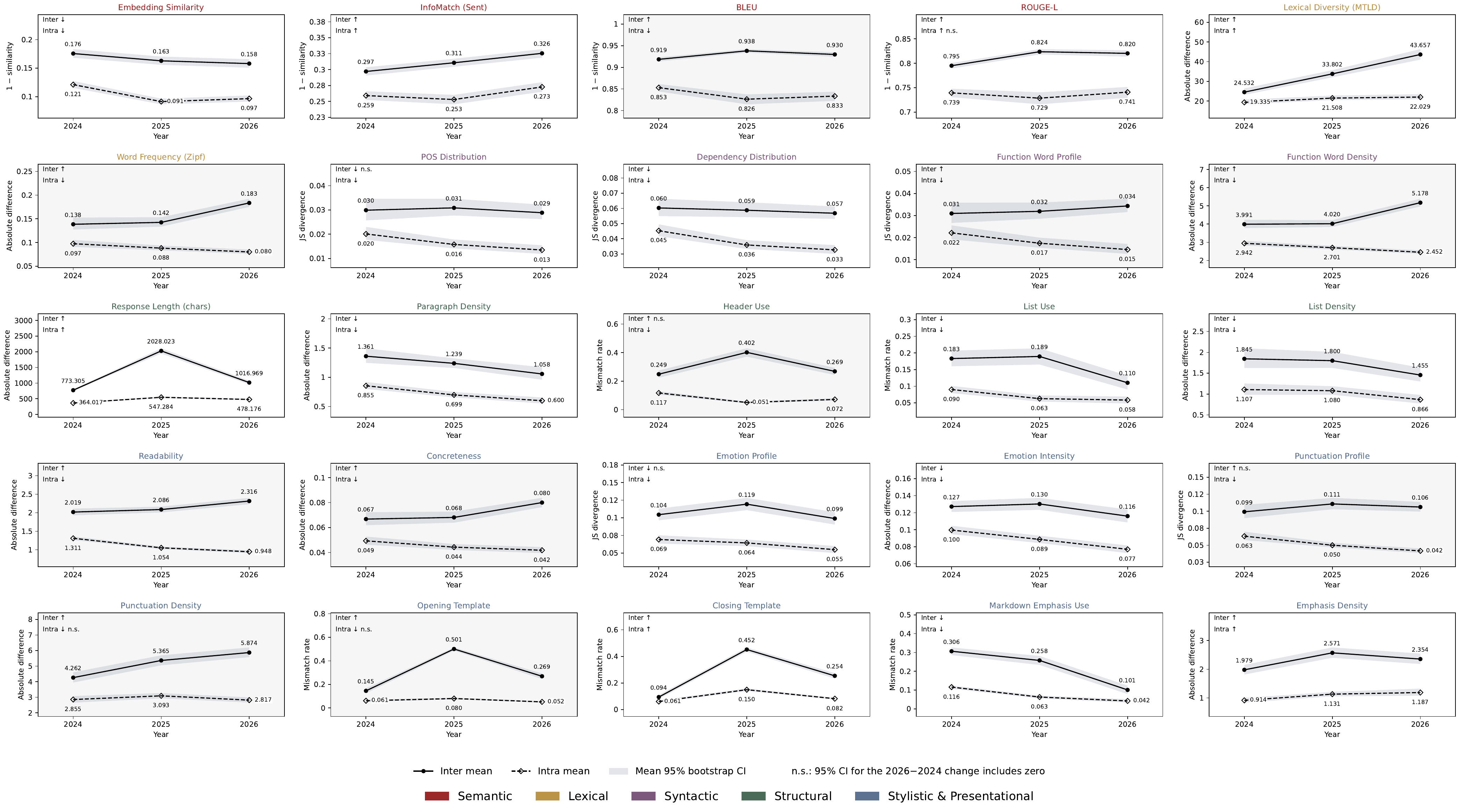}
    \caption{Inter- and intra-model distances across all 25 textual metrics from 2024 to 2026. Shaded bands indicate 95\% bootstrap confidence intervals (10,000 prompt-level resamples, with prompts jointly resampled across models and years). Gray backgrounds indicate metrics with increasing inter-model and decreasing intra-model distances, and arrows indicate the direction of change from 2024 to 2026. Lower values indicate greater similarity.}
    \label{fig:temporal_trends}
\end{figure*}

\begin{figure*}[!t]
    \centering
    \includegraphics[width=\textwidth]{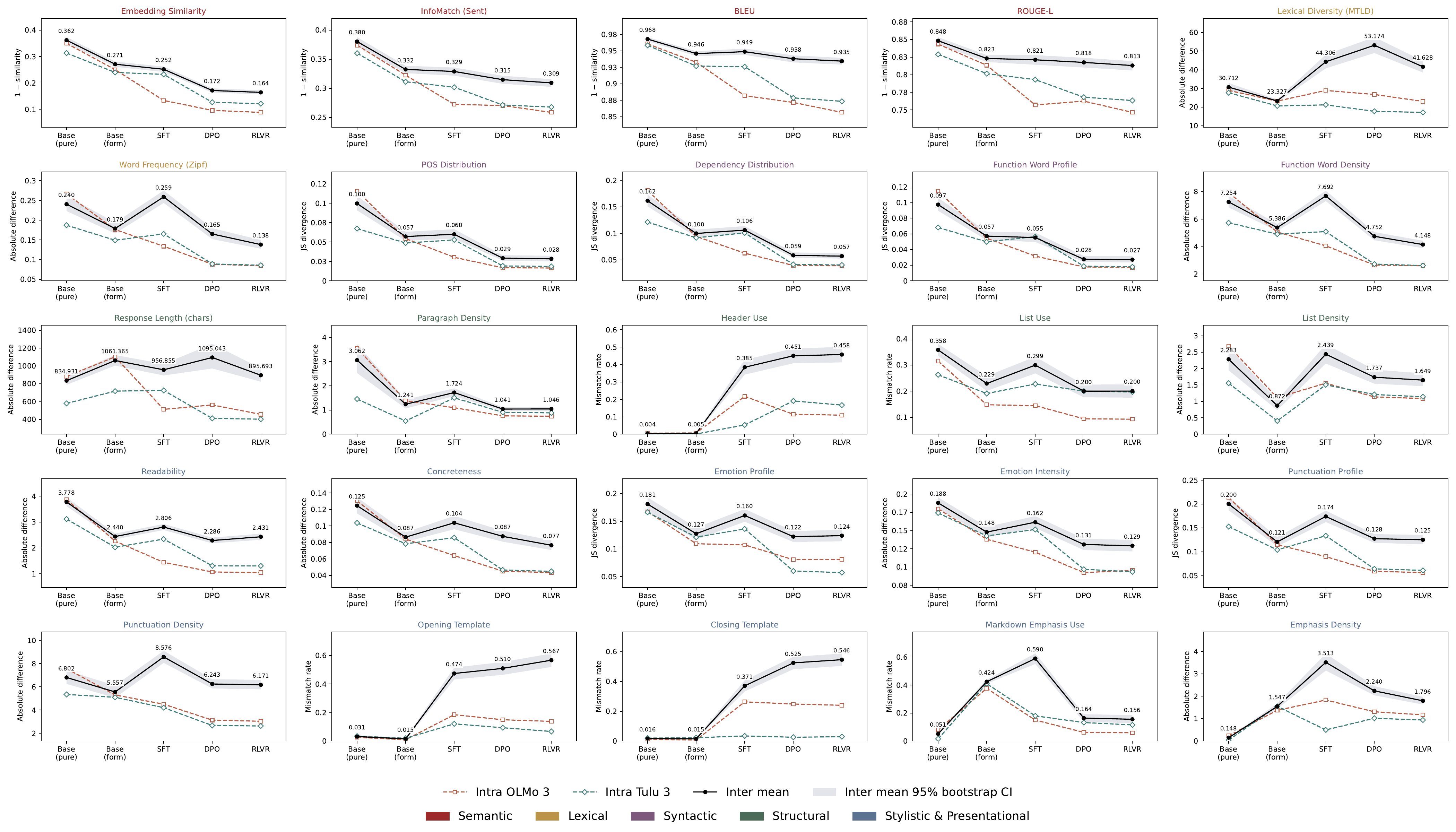}
    \caption{Inter- and intra-model distances across all 25 textual metrics for OLMo~3 and T\"ulu~3 over post-training stages. Solid black lines denote inter-model distances; dashed orange and blue lines denote intra-model distances for OLMo~3 and T\"ulu~3, respectively. Shaded bands indicate 95\% bootstrap confidence intervals (10,000 prompt-level resamples). Base (\textit{pure}/\textit{form}) denotes two alternative few-shot prompting conditions, and SFT, DPO, and RLVR denote successive post-training checkpoints. Lower values indicate greater similarity.}
    \label{fig:stage_full}
\end{figure*}

We present the complete inter- and intra-model distance trajectories across metrics. Figure~\ref{fig:temporal_trends} shows the temporal evolution from 2024 to 2026, and Figure~\ref{fig:stage_full} shows the evolution across post-training stages for OLMo~3 and T\"ulu~3 (Section~\ref{sec:stage}).

\subsection{Similarity Across Question Types}
Figure~\ref{fig:question_type} reports inter-model similarity across six question categories. The overall pattern is consistent across categories: embedding similarity and sentence-level InfoMatch remain relatively high, while lexical overlap measured by BLEU and ROUGE-L is substantially lower. Some variation exists across question types. In particular, Alternative Styles exhibits the highest InfoMatch, BLEU, and ROUGE-L scores, suggesting greater overlap when the writing objective is more constrained. Overall, these results indicate that the distinction between shared semantic content and diverse lexical realizations holds across different question types.

\begin{table}[H]
\centering
\scriptsize
\begin{tabular}{lccc}
\toprule
Metric & Qwen3 & Jina & QZhou \\
\midrule
\multicolumn{4}{l}{\textit{Semantic}} \\
\quad InfoMatch (Sent)$^\dagger$ & 0.646$^{***}$ & 0.725$^{***}$ & 0.599$^{***}$ \\
\quad BLEU & 0.633$^{***}$ & 0.559$^{***}$ & 0.482$^{***}$ \\
\quad ROUGE-L & 0.636$^{***}$ & 0.537$^{***}$ & 0.468$^{***}$ \\
\midrule
\multicolumn{4}{l}{\textit{Lexical}} \\
\quad Lexical Diversity (MTLD) & 0.019 & $-$0.053 & $-$0.022 \\
\quad Word Frequency (Zipf) & 0.262$^{***}$ & 0.287$^{***}$ & 0.249$^{***}$ \\
\midrule
\multicolumn{4}{l}{\textit{Syntactic}} \\
\quad POS Distribution & 0.535$^{***}$ & 0.621$^{***}$ & 0.483$^{***}$ \\
\quad Dependency Distribution & 0.529$^{***}$ & 0.621$^{***}$ & 0.526$^{***}$ \\
\quad Function Word Profile & 0.389$^{***}$ & 0.453$^{***}$ & 0.350$^{***}$ \\
\quad Function Word Density & 0.236$^{***}$ & 0.258$^{***}$ & 0.211$^{***}$ \\
\midrule
\multicolumn{4}{l}{\textit{Structural}} \\
\quad Response Length (chars) & 0.075 & $-$0.111$^{*}$ & $-$0.021 \\
\quad Paragraph Density & 0.392$^{***}$ & 0.414$^{***}$ & 0.321$^{***}$ \\
\quad Header Use & $-$0.020 & $-$0.016 & $-$0.032 \\
\quad List Use & 0.139$^{**}$ & 0.057 & 0.065 \\
\quad List Density & 0.177$^{***}$ & 0.074 & 0.050 \\
\midrule
\multicolumn{4}{l}{\textit{Stylistic \& Presentational}} \\
\quad Readability & 0.206$^{***}$ & 0.199$^{***}$ & 0.245$^{***}$ \\
\quad Concreteness & 0.240$^{***}$ & 0.313$^{***}$ & 0.254$^{***}$ \\
\quad Emotion Profile & 0.323$^{***}$ & 0.350$^{***}$ & 0.336$^{***}$ \\
\quad Emotion Intensity & 0.209$^{***}$ & 0.231$^{***}$ & 0.212$^{***}$ \\
\quad Punctuation Profile & 0.348$^{***}$ & 0.365$^{***}$ & 0.332$^{***}$ \\
\quad Punctuation Density & 0.323$^{***}$ & 0.382$^{***}$ & 0.316$^{***}$ \\
\quad Opening Template & $-$0.036 & $-$0.060 & $-$0.009 \\
\quad Closing Template & 0.108$^{*}$ & 0.112$^{*}$ & 0.081 \\
\quad Markdown Emphasis Use & 0.163$^{***}$ & 0.098$^{*}$ & 0.104$^{*}$ \\
\quad Emphasis Density & 0.188$^{***}$ & 0.135$^{**}$ & 0.107$^{*}$ \\
\bottomrule
\end{tabular}
\caption{Spearman correlations between whole-response embedding similarity and each response-similarity metric on \textsc{Infinity-Chat}, using Qwen3-Embedding-8B, jina-embeddings-v5-text, and QZhou-Embedding ($n=500$ response pairs). For each metric, pairs that do not meet its computation requirements are excluded. $^{*}p<0.05$, $^{**}p<0.01$, $^{***}p<0.001$. $^\dagger$InfoMatch is computed with the same embedding model as the corresponding column.}
\label{tab:embedding_robustness}
\end{table}

\end{document}